\documentclass{llncs}

\usepackage[rgb,dvipsnames]{xcolor}
\usepackage{graphicx}

\usepackage{tikz}
\usepackage{calc}
\usepackage{amsmath,amssymb,mathtools} \usepackage{color}
\usepackage{etoolbox}
\usepackage{dsfont}
\usepackage[inline,shortlabels]{enumitem}
\usepackage{multirow,booktabs}
\usepackage{adjustbox}
\usepackage{placeins}
\usepackage{altcopyrightbox}
\usepackage{wrapfig}
\usepackage{soul}

\usepackage{fancyhdr}
\fancypagestyle{style1}{
\fancyhf{}
\fancyhead[LO]{Spatial Consistency Loss \hfill \thepage}
\fancyhead[LE]{\thepage \hfill T.\ Verelst, P.\ K.\ Rubenstein, M.\ Eichner, T.\ Tuytelaars, M.\ Berman}
\fancyfoot[CE,CO]{}
\fancyfoot[LE,RO]{}
}
\fancypagestyle{style2}{
\fancyhf{}
\fancyhead[LO]{Spatial Consistency Loss -- Supplementary\hfill \thepage}
\fancyhead[LE]{\thepage \hfill T.\ Verelst, P.\ K.\ Rubenstein, M.\ Eichner, T.\ Tuytelaars, M.\ Berman}
\fancyfoot[CE,CO]{}
\fancyfoot[LE,RO]{}
}

\usepackage[binary-units]{siunitx}
\usepackage{etoolbox}
\robustify\bfseries

\usepackage{chngcntr}  

\usepackage{arydshln}  \newcolumntype{H}{>{\setbox0=\hbox\bgroup}c<{\egroup}@{}}  

\usepackage{ccicons}

\usepackage{hyperref}
\usepackage[]{cleveref}  

\makeatletter
\renewcommand{\CRB@setcopyrightfont}{\tiny
\color{black!33}
\sffamily
}
\newlength{\cheight}
\newcommand{\fullwidthcopyrightimage}[2]{\settoheight{\cheight}{\hbox{\tiny #1}}\copyrightbox[r]{\includegraphics[width=\linewidth - 1ex - \cheight]{#1}}{#2}}

\newcommand{\licenseCCBy}{\href{https://creativecommons.org/licenses/by-nc/2.0/}{\ccLogo \ccAttribution \ccNonCommercial}}

\newcommand*{\affmark}[1][*]{\textsuperscript{#1}}

\usepackage[caption=false]{subfig}

\usepackage{xspace}
\newcommand*{\eg}{e.g.\@\xspace}

\DeclareMathOperator*{\argmax}{argmax}

\newcommand*{\eqsp}{\,}

\DeclarePairedDelimiter{\norm}{\lVert}{\rVert}

\newcommand{\missing}{\emptyset}

\makeatletter
\newcommand*{\etc}{\@ifnextchar{.}{etc}{etc.\@\xspace}}
\newcommand*{\etal}{\@ifnextchar{.}{et al}{et al.\@\xspace}}
\makeatother

\begin{document}
\mainmatter

\title{Spatial Consistency Loss for Training Multi-Label Classifiers from Single-Label Annotations}

\titlerunning{Spatial Consistency Loss} 
\authorrunning{T.\ Verelst, P.\ K.\ Rubenstein, M.\ Eichner, T.\ Tuytelaars, M.\ Berman} 
\author{Thomas Verelst\affmark[1]\thanks{Work done during an internship at Apple.}
\quad Paul K.\ Rubenstein\affmark[2]\quad Marcin Eichner\affmark[2]\\ Tinne Tuytelaars\affmark[1]\quad Maxim Berman\affmark[2]}
\institute{\affmark[1]ESAT-PSI, KU Leuven, Belgium \quad \affmark[2]Apple\vspace{-1em}}

\maketitle

\begin{abstract}
As natural images usually contain multiple objects, multi-label image classification is more applicable ``in the wild'' than single-label classification. However, exhaustively annotating images with every object of interest is costly and time-consuming. We aim to train multi-label classifiers from single-label annotations only. We show that adding a consistency loss, ensuring that the predictions of the network are consistent over consecutive training epochs, is a simple yet effective method to train multi-label classifiers in a weakly supervised setting. We further extend this approach spatially, by ensuring consistency of the spatial feature maps produced over consecutive training epochs, maintaining per-class running-average heatmaps for each training image. We show that this spatial consistency loss further improves the multi-label mAP of the classifiers. In addition, we show that this method overcomes shortcomings of the ``crop'' data-augmentation by recovering correct supervision signal even when most of the single ground truth object is cropped out of the input image by the data augmentation. We demonstrate gains of the consistency and spatial consistency losses over the binary cross-entropy baseline, and over competing methods, on MS-COCO and Pascal VOC. We also demonstrate improved multi-label classification mAP on ImageNet-1K using the ReaL multi-label validation set.\vspace{-0.5em}
\end{abstract}
 \section{Introduction}

In the last decade, computer vision has seen great progress thanks to the emergence of large-scale data-driven machine learning. 
With enough annotated data, machine perception has reached or exceeded human accuracy in many difficult tasks, in particular single-label image classification \cite{ILSVRC15}. 
Yet obtaining large amounts of annotated data remains a challenge, especially in more granular object recognition tasks such as multi-label classification, object detection or instance segmentation. 
Exhaustively annotating all objects in images on a large scale is tedious, time-consuming, and error prone. 
In some cases, such as in the medical domain, recruiting enough domain experts for producing refined annotations on millions of images may be infeasible in practice. 

\begin{figure}[t]
    \centering
    \fullwidthcopyrightimage{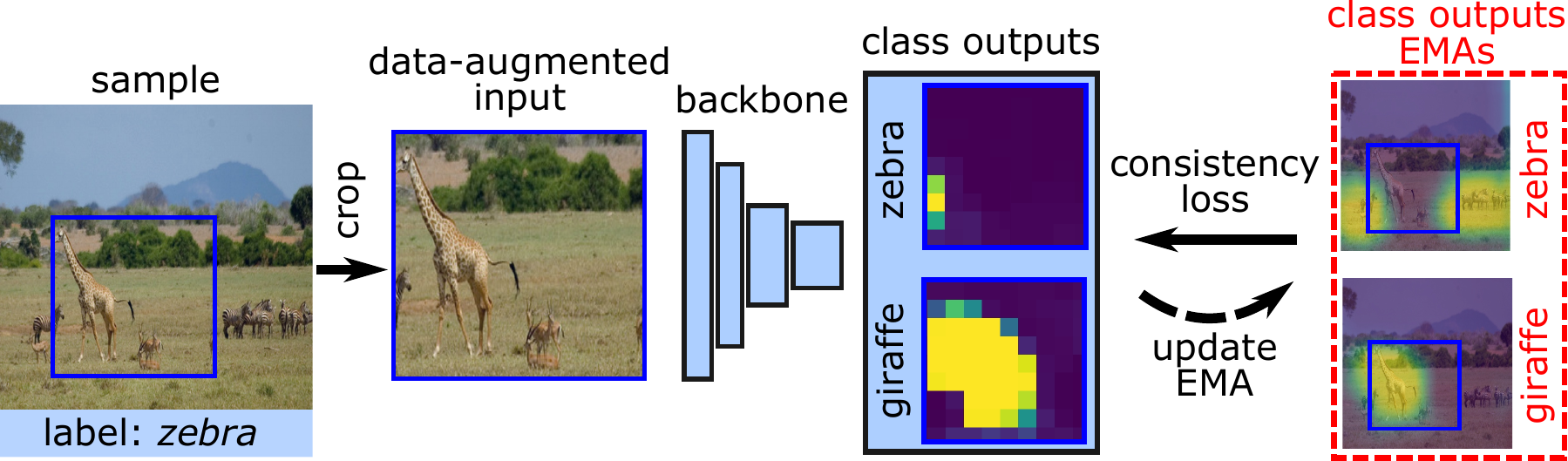}{\href{https://www.flickr.com/photos/lathos/3953904506/}{Photo by S.\ Cozens} \href{https://creativecommons.org/licenses/by-nc/2.0/}{\ccLogo \ccAttribution \ccNonCommercial}}
    \caption{{
    We train a multi-label classifier from a dataset of single-label images. 
In this example, only the \emph{zebra} is annotated.  After the random data-augmentation crop, the zebra are no longer visible -- the single label no longer matches the image input. 
    The spatial consistency loss ensures consistency between 
    \begin{enumerate*}[(i)]
    \protect\item the network's output classification maps \protect\item exponential moving averages (EMAs) of these output maps over successive training epochs.
    \end{enumerate*}
These spatial averages alleviate the adversarial impact of objects absent from the ground truth during training (\emph{giraffe} in this example).
    }}
    \label{fig:splash}
\end{figure} 
To reduce the annotation cost, some large-scale datasets such as OpenImages~\cite{kuznetsova2020openimages} only annotate a subset of the object classes for each image in the dataset.
In this case, the annotation process yields a set of positive labels guaranteed to be in the image, a set of negative labels guaranteed to be absent from the image, and a set of unknown labels for which no information is provided.

A more extreme setting, which reduces the annotation effort substantially, is the annotation of a single positive label per image, with no negative labels. In particular, ImageNet~\cite{ILSVRC15} is annotated with a single label per image, intended to represent the main object of interest. 
While this type of annotation is sensible for a single label classification task, it is clear that most natural images contain more than one object. 
This can cause some ambiguity as the problem of classifying an image containing multiple objects into a single category is not well posed. 
Therefore, using single labels for scenes containing multiple objects introduces a specific type of label noise, which can hurt the performance of the classifier. 
Most contemporary deep learning approaches for large scale single-label classification are trained using a 
``one-versus-all'' cross-entropy loss, where the target distribution $P(s|\mathbf{x}_n)$ of predicted object class $s$ on training sample $(\mathbf{x}_n, s_n)$ is a Dirac distribution $\mathds{1}_{[s=s_n]}$. 
This modeling assumes that the image represents a single object only. 
To some extent, the use of a top-$k$ evaluation metric can mitigate this issue at evaluation time, allowing the classifier to return multiple candidate classes to match the ground truth~\cite{ILSVRC15}. 
However, this in an imperfect solution, which bypasses the problem rather than solving it.

Regularization, either implicitly through \eg{} stochastic optimization or explicitly through the use of label smoothing techniques~\cite{szegedy2016rethinking,Wightman2021ResNetSB}, can improve the accuracy and help in making the classifier learn a useful mapping in spite of the inherent label noise. 
Other lines of work aim to go beyond regularization, and acknowledge that the images of single-label datasets such as ImageNet can contain more than one object in practice~\cite{recht2019imagenet,beyer2020we_real,yun2021relabel_imagenet}. 
In such a setting, a single-labeled dataset can be thought of as a weakly-labeled multi-label classification dataset, with a single positive label annotated per image.

When training neural network classifiers, previous works show that one can leverage additional unannotated data by ensuring consistency of the network outputs on this data among consecutive training epochs~\cite{temporal_ensembling}. 
In this work, we first show that a similar approach, easy to implement, is already competitive with existing works in the single-positive label setting. 
To this end, we keep exponential moving averages (EMAs) of the network outputs over the training epochs for each training sample, and add a consistency loss (CL) to favor consistency between the network outputs and these running averages.

We extend this approach in the spatial domain. 
In agreement with existing work~\cite{yun2021relabel_imagenet}, we observe
that the single-positive annotation is particularly detrimental to the training of high-accuracy classifiers when used in conjunction with image crops as a data-augmentation technique. 
While random cropping is used in the training of state of the art 
image classifiers~\cite{krizhevsky2012imagenet,tan_efficientdet_2020},
cropping an image risks removing the object corresponding to the ground truth annotation, leading the optimization to be misguided by the ground truth label, as in \cref{fig:splash}. 
We introduce a spatial consistency loss (SCL) to further mitigate multi-object label noise in single-labeled datasets, and address the label discrepancy introduced by cropping data-augmentation. 
By taking EMAs of the spatial outputs of the network over consecutive training epochs, we obtain spatial heatmaps which localize objects in the image, beyond the single ground truth label. 
The SCL uses these spatial running averages as additional source of self-supervision which further improves the final multi-label classification accuracy of the network.

The contributions of this work are as follows:
\begin{itemize} 
    \setlength{\itemsep}{0pt}
    \setlength{\parskip}{0pt}
    \setlength{\parsep}{0pt}
    \item We show that consistency losses yield competitive results for training multi-label classifiers from single positive label annotations;
    \item We introduce a spatial consistency loss, and show that it can improve the multi-label accuracy, and acts in synergy with the ubiquitous ``resize+crop'' data-augmentation;
    \item We propose a scheme to mine labels expected to be unannotated positives and ignore them in the loss, and show that ignoring these expected positives is essential for good performance.
    \item We show that our contributions yield improvements on the MS-COCO and Pascal VOC in the single positive setting, as well as on ImageNet-1K evaluated using multi-label annotations.
\end{itemize} \section{Related work}

\paragraph{Partial annotations.} Collecting exhaustive multi-label classification annotations on a large number of classes and images can be intractable, which is why many large-scale datasets resort to partial annotations~\cite{liu2021emerging}.
For instance, for each image in OpenImages~\cite{kuznetsova2020openimages} and LVIS~\cite{Gupta2019LVISAD}, only a small fraction of the labels are annotated as positives or negatives. 
Collecting a larger amount of partially labeled data can sometimes lead to better performance than a smaller set of fully-annotated data~\cite{durand_learning_2019}. 
Partial labels can also occur naturally when training a model on the combination of several datasets with disjoint label spaces~\cite{yan_learning_2020,zhao2020object}.

Multi-label learning with missing labels can be framed as a transductive learning problem, where one aims to explicitly recover complete annotations that are consistent with the partial annotations provided~\cite{wu_multi-label_2014}. 
Graph neural networks~\cite{wu_multi-label_2018,chen2019multi_graph,durand_learning_2019,wang_multi-label_2019,lyu_partial_2020,huynh_interactive_2020,li2020learning} or adversarial training~\cite{yan2021adversarial} can be used to predict the missing labels based on the annotated ones. 
The label co-occurrence structure can also be leveraged to estimate the confidence of labels~\cite{bi_multilabel_2014,ibrahim_confidence-based_2020}.

A simple way to optimize with missing labels is to treat them as negatives~\cite{sun2010multi_weak_label,bucak2011multi}. 
However, unannotated positives can deteriorate performance. 
Ignoring the unannotated classes in the loss function can alleviate this issue~\cite{durand_learning_2019}, but this is inapplicable when the annotations only contain positives~\cite{cole2021multi}. 
Asymmetric loss design can help handle missing labels beyond the BCE loss~\cite{zhang2021simple_assymetric}. 

Prior works consider single-positive labels in the single-label setting~\cite{cole2021multi,zhang2021simple_assymetric}, as a combination of single-label learning~\cite{qiu2017nonconvex_singlelabel,duan2019learning_singlelabel,hu2020one} and positive-unlabeled learning~\cite{du2014analysis_positiveunlabeled,bekker2020learning_positiveonly}. 
\cite{cole2021multi} propose to go beyond label smoothing~\cite{szegedy2016rethinking,Wightman2021ResNetSB} to deal with the label noise introduced by false negative labels their regularized online label estimation (ROLE) method estimates the missing labels in an online fashion, by jointly optimizing a label estimator and image classifier, the output of the former serving as ground truth of the latter. 
We also propose a method to mine annotations expected to be positives in \cref{sec:expected-negative} by keeping running averages of the class specific scores on the training set images, and show that this simpler method performs similarly to ROLE in~\cref{sec:experiments}.

\paragraph{Semi-supervised learning.}
Semi-supervised learning uses a set of unlabeled data samples in addition to the fully-labeled samples, and can be seen as a special case of partial annotation~\cite{guo2012semi}. 
One way to incorporate unlabeled samples in the training process is by encouraging consistency of predictions on these samples over different epochs or augmentations. 
Ladder networks~\cite{rasmus2015semi} encourage consistency between a standard branch and the denoised predictions of a corrupted branch. 
\cite{temporal_ensembling} proposes the $\mathrm{\Pi}$-model, enforcing consistency between two perturbed versions of the same sample. 
In addition, they propose self-ensembling to build a consensus prediction by averaging outputs among different training epochs.
Our consistency losses in \cref{sec:cl,sec:scl} applies similar ideas directly on the training set, rather than a held-out dataset of unlabeled images.

Other methods use pseudo-labeling to leverage unannotated images. 
\cite{lee2013pseudo} uses the highest-scoring class as the true label for unlabeled data. 
However, pseudo labels are prone to concept drift and confirmation bias, where early mislabeled samples lead to accumulating errors. 
Curriculum labeling~\cite{cascante2021curriculum} mitigates this using a refined training strategy.
Noisy student~\cite{xie_self-training_2020} demonstrated state of the art results on ImageNet~\cite{krizhevsky2012imagenet} using self-training and distillation on a large set of unlabeled images, by iterative relabeling data and using increasingly larger student models. 
By contrast, we choose to ignore the labels that we identify as possible positives in~\cref{sec:expected-negative} rather than incorporating them in the positive annotations, avoiding the concept drift issue which we also identify in our comparison of \cref{sec:abl}.

\paragraph{Data augmentation and instance discrimination.} Our CL and SCL losses enforce consistency of the network across subsequent training epochs. 
As the network will see different data-augmentations of the training samples in different epochs, this also favors invariance of the network outputs to the data-augmentation.
This can be connected to recent trends of self-supervised learning which focus on the instance discrimination task, and ensure that the embeddings of data-augmented versions of an instance should be closer in embedding space than the embeddings of different instances~\cite{van2018representation,misra2020self,he2020momentum,henaff2020data,chen2020simple}, an approach which has also demonstrated to improve classification performance in the supervised setting~\cite{Khosla2020SupCon}. 
In the fully annotated multi-label image classification setting,~\cite{guo2019visual} encourages consistency of the spatial activations of the network among two data augmentations of an image, akin to a spatial extension of the $\mathrm{\Pi}$-model of \cite{temporal_ensembling}. 
In the semi-supervised single-label setting, our SCL of \cref{sec:scl} uses a similar idea of encouraging consistency of the spatial network class outputs, but uses a temporal ensemble over the different training epochs to do so, rather than directly comparing the outputs of data-augmented copies during a single training iteration.

 \section{Method}

\subsection{Problem statement}
We state the problem of multi-label classification with partially annotated labels similarly to \cite{cole2021multi}. 
Our goal is to learn a mapping from an image $\mathbf{x}_n$ to the indicator vector $\mathbf{y}_n \in \{0,1\}^L$ of the classes contained in the image, $L$ being the number of classes. 
We use a dataset $(\mathbf{x}_n, \mathbf{z}_n)_{n=1}^N$, where each input image $\mathbf{x}_n$ has a partial annotation $\mathbf{z}_n \in \{0,1,\missing\}^L$. 
The positive labels encoded by $1$ are contained in the image; the negative labels $0$ are absent from the image; missing labels encoded by $\missing$ can be either present or absent. In the single positive setting, there is a single positive label $i$ for each image such that $z_{ni} = 1$; all other labels $j \neq i$ are supposed unknown ($z_{nj} = \missing$).

Given an image $\mathbf{x}_n$, a neural network classifier predicts $L$
label probabilities 
$\mathbf{f}_n \in [0, 1]^L$. At training time, the network parameters are optimized to minimize the empirical risk on the training set, measured with a loss function~$\mathcal{L}$. 
A common multi-label classification loss is the binary cross entropy (BCE) loss \begin{equation}
\label{eq:bceloss}
    \mathcal{L}_{\text{BCE}}(\mathbf{f}_n) =  -\frac{1}{L} \sum^{L}_{i=1}\, [z_{ni}=1] \log(f_{ni}) +
    [z_{ni}=0]  \log(1-f_{ni})
\end{equation}
with $[\cdot]\in\{0, 1\}$ the Iverson bracket equal to $1$ iff.\ the condition holds. With incomplete annotations, missing labels where $z_{ni} = \missing$ are ignored in \cref{eq:bceloss} and thus not penalized. 
Although natural, this modeling is not suited for training with only positive annotated labels, such as the single positive setting that we consider.  
Indeed, in such a setting, nothing prevents the network from predicting all $L$ classes regardless of the input, as there is no penalty for false positives.

\subsection{Assume-negative loss (AN)}
One simple strategy to handle single-positive labels is to assume that all unknown labels are negatives.  
This leads to the assume negative (AN) loss function~\cite{cole2021multi}\begin{equation} 
\label{eq:anloss}
    \mathcal{L}_{\text{AN}}(\mathbf{f}_n) =  -\frac{1}{L} \sum^{L}_{i=1}\, [z_{ni}=1] \log(f_{ni})
    +
    [z_{ni} \in \{0,\missing\}]  \log(1-f_{ni})\eqsp.
\end{equation}
In this case, unobserved labels where $z_{ni} = \missing$ are considered as negatives. 
This is justifiable since the number of objects present in an image is typically small, leading to only a few false negatives in the supervision, 
weighed against many true negatives supervised correctly. 
However, the false negatives of the AN loss can have a large impact on the accuracy. 
Our experiments in \cref{sec:biasANSP} show that the AN loss can steer the network towards predicting only a single positive label per image, even in the presence of multiple objects.

\subsection{Consistency loss (CL) \label{sec:cl}}
We show that adding a consistency loss, a strategy commonly used in semi-supervised methods with unannotated samples~\cite{temporal_ensembling,tarvainen2017mean}, yields competitive results when training classifiers with single-positive labels. 
Over consecutive training epochs, the network sees different data-augmented versions of an image; keeping running averages of the model outputs on these different augmentations
leads to more robust label estimates, which we use as supervision. At training epoch $t$, the estimated scores $\mathbf{s}_{n}^t$ are updated with the network outputs $\mathbf{f}_n^t$ as an EMA
\begin{equation}
\label{eq:runningestimate}
    \mathbf{s}_{n}^t = \mu\ \mathbf{s}_{n}^{t - 1} + (1-\mu)\mathbf{f}_{n}^t
\end{equation}
with $\mu$ the momentum.
The scores $\mathbf{s}_n^{0}$ are initialized to 1 for the positive label, i.e. $s_{ni}^{0} = 1$  if $z_{ni} = 1$, and $0$ otherwise.
The consistency loss (CL) is given by the $\ell_2$-distance between the predicted scores and the running-average estimates:
\begin{equation}
\label{eq:consistency_loss}
    \mathcal{L}_{\text{CL}}(\mathbf{f}_n^t) =  \norm{\mathbf{f}_n^t - \mathbf{s}_n^{t-1}}_2\eqsp.
\end{equation}

\subsection{Spatial consistency loss (SCL) \label{sec:scl}}
We extend the CL to the spatial dimensions, by predicting class scores for each spatial coordinate of the feature map. 
This method applies to a typical classifier network architecture with a convolutional backbone, an average pooling operation over the features and a fully connected classification layer. 
To obtain spatially localized class-specific predictions, we modify the network architecture by \begin{enumerate*}[label=(\roman*)]\item interpreting the fully connected layer as a $1{\times}1$ convolution, and \item applying it before the pooling operation rather than after. 
\end{enumerate*}
Assuming square input images for the sake of exposition, this modification produces spatial score maps $\mathbf{F}_n \in [0,1]^{G \times G \times L}$, with $G{\times}G$ the spatial dimensions of the feature map. 
Applying the 
fully-connected layer to every spatial location of the feature map increases the computations at training time. 
However, due to the distributive property, the order of the average pooling and the $1\times 1$ convolution layers can be reversed  
without affecting the network outputs. 
Consequently, our modification causes no computational penalty during inference.

For each image $n$, we keep score heatmaps $\mathbf{H}_n^t \in [0,1]^{W \times W \times L}$ which contain running averages of the output score maps $\mathbf{F}_n^t$ at epoch $t$. The heatmap size $W$ is a multiple of $G$, allowing to store details in the heatmaps at a finer resolution than the score maps; in practice, we use $W = 2G$.
When feeding the input $\mathbf{x}_n$ to the network, 
we record the spatial transformation $T_n^t$ used in the data augmentation, such as cropping and flipping. 
Given this transformation, only the visible part of the heatmaps $\mathbf{H}_n^t$ is updated with an EMA: 
the score maps $\mathbf{F}_n^t$ are resized with bilinear interpolation to fit the cropped region, and flipped if needed. Heatmap regions that are cropped out of the input are not updated.
As for the CL method, the heatmaps are initialized to $1$ for the annotated ground truth and $0$ for the other classes.

The spatial consistency loss (SCL) is given by the $\ell_2$-norm between the score heatmap and the network output. The input augmentation transformation $T_n^t$ is first applied on the running-average heatmap. The result is then rescaled to match the dimensions of $\mathbf{F}_n^t$. The spatial consistency loss is given by
\begin{equation}
\label{eq:spatial_consistency_loss}
    \mathcal{L}_{\text{SCL}}(\mathbf{F}_n^t) =  \| \mathbf{F}_n^t -  \text{resize}(T_n^t(\mathbf{H}_n^{t-1})) \|_2\eqsp.
\end{equation}

\subsection{Expected-negative loss (EN) \label{sec:expected-negative}}
We use the CL and SCL loss detailed in \cref{sec:scl,sec:cl} in conjunction with a loss based on the BCE (\cref{eq:bceloss}). 
While we can use the AN loss of \cref{eq:anloss}, our experiments in \cref{sec:abl} show that 
the bias of the AN loss towards negative labels impacts the network's ability to predict positives. 
Our interpretation is that the network is penalized strongly by the BCE when predicting high scores for missing positive labels. 
Therefore, the missing positives labels in AN lead to a large incorrect supervision that can dominate the contribution to the loss from the true negatives. 

We design a strategy to ignore these large incorrect contributions to the loss, by tracking a set of samples that we expect to be positives for each class. 
To this effect, we leverage the running-average scores $\mathbf{s}^t_n \in [0,1]^L$ computed according to \cref{eq:runningestimate} by the CL method. 
We also compute these score estimates when using SCL, with minimal overhead compared to the storage of the heatmaps. 
We use a hyper-parameter $K$ which sets the number of expected positives labels per image. 
For a training set of size $N$, the expected number of ground truth positives with class $i$ is given by
\begin{equation}
\label{eq:expected_num_pseudo_pos}
p_i = K N \cdot \frac{\sum_{n=1}^N{[z_{ni} = 1]}}{N} =  K\sum_{n=1}^N{[z_{ni} = 1]}\eqsp,
\end{equation}
assuming that the class distribution of annotated labels ${\sum_{n=1}^N{[z_{ni} = 1]}}/{N}$ is similar to the unknown true distribution ${\sum_{n=1}^N{y_{ni}}}/{N}$.
At the end of each epoch $t$, we identify the top-$p_i$ instances for each class $i$ among the running-average score estimates $(s^t_{ni})_{n=1\ldots N}$ as likely to correspond to positive ground-truth labels. 
We set $\hat{z}_{ni}^t \in \{0, 1\}$ as indicator vectors of these expected positive labels. 
In the first training epoch, we initialize $\hat{z}_{ni}^0 = 1$ if ${z}_{ni} = 1$ and 0 otherwise. 
Finally, we modify the AN loss by ignoring the labels that are among the expected positives in the computation of the loss. 
This yields the expected negative (EN) loss
\begin{equation} 
\label{eq:loss_ignore}
    \mathcal{L}_{\text{EN}}(\mathbf{f}_n) =  -\frac{1}{L} \sum^{L}_{i=1}\, [z_{ni}=1] \log(f_{ni}) +
    [\hat{z}^t_{ni}=0]  \log(1-f_{ni})\eqsp.
\end{equation}
Contrary to the AN loss, $\mathcal{L}_{\text{EN}}$ does not assume all unannotated labels to be negatives, but only the ones that are not part of the expected positive samples. 

In our experiments, we use the EN loss in combination with the CL or SCL
\begin{equation}\label{eq:final_loss}
    \mathcal{L} = \mathcal{L}_{\text{EN}} + \gamma\mathcal{L}_{\text{(S)CL}}
\end{equation}
with $\gamma$ a weighting parameter.

 \section{Experiments \label{sec:experiments}}

\subsection{Comparison on MS-COCO and Pascal VOC}

\subsubsection{Dataset, setup and metrics.}
\label{sec:setup}
We use MS-COCO 2014~\cite{lin2014microsoft_coco} and Pascal VOC 2012~\cite{pascal-voc-2012} as benchmarks for multi-label classification. 
In order to test our contributions, we use the code shared by~\cite{cole2021multi} to simulate a single-positive annotated setting, and reproduce their train, validation and test samples. 
In this setup, 20\% of the original train data is first set aside for validation. 
For the remaining training samples, a single label is picked at random among the ground truth labels to be used as single-positive annotation during training. 
The original official multi-label validation sets are used as test splits. 
Consequently, we obtain partially annotated training samples, and fully-annotated validation and testing samples. 
The train/val/test splits have size 64K/16K/40K on MS-COCO, and 4.6K/1.1K/5.8K images on Pascal VOC.

We perform experiments with the ResNet-50~\cite{he2016deep_resnet} model provided by torchvision~\cite{NEURIPS2019_9015}, and train with input images of size $448{\times}448$, as in~\cite{cole2021multi}. 
We use random crop augmentations (area scale 0.25 to 1) and random horizontal flip; additional details and ablation on the scale are provided in \cref{suppl:dataaug}.
We use the Adam optimizer~\cite{KingmaB14Adam} with a batch size of 8. 
When trained from scratch without pretraining, the model is trained for 100 epochs with learning rate $10^{-4}$ and cosine learning rate decay. 
With ImageNet-1k pretraining~\cite{ILSVRC15}, the final linear layer is trained for 5 epochs, followed by 25 epochs of finetuning of the whole network with a learning rate of $10^{-5}$. 

We report the mean average precision (mAP) on the test split, using the epoch corresponding to the best validation mAP. 
We note that ROLE~\cite{cole2021multi} report the test value corresponding to the best validation among 6 combinations of learning rates and batch sizes for each experiment.
For a fair comparison, we experiment with the codebase shared by~\cite{cole2021multi} to report the performance of ROLE under the same training and evaluation setup. 

\begin{table*}[t]
\centering
\caption{Mean average precision (mAP) obtained on the test set of Pascal VOC 2012~\cite{pascal-voc-2012} and MS-COCO 2014~\cite{lin2014microsoft_coco}, both with and without pretraining on ImageNet-1K~\cite{ILSVRC15}. 
Results indicated with $\dag$ are reported by related work. 
}

\label{tab:results}
\centering
\adjustbox{max width=\textwidth}{\begin{tabular}{@{}lccccc@{}}
\toprule
\multirow{2}{*}{\textbf{Method}}         & \multirow{2}{*}{\textbf{Supervision}}& \multicolumn{2}{c}{\textbf{No pretraining}}                  & \multicolumn{2}{c}{\textbf{IN1K pretraining}}  \\ \cmidrule(lr){3-4} \cmidrule(lr){5-6}
                                         &                                      & \scriptsize VOC12                          & \scriptsize MS-COCO& \scriptsize VOC12                            & \scriptsize MS-COCO          \\ \midrule
fully-annotated oracle (BCE)                   & all pos \& neg                       & 53.4                                       & 64.8            & 90.7                                         & 79.3    \\ \midrule
AN + label smoothing \cite{cole2021multi}& 1 pos / img                          & -                                          & -               & 86.5$^\dag$                                  & 69.2$^\dag$    \\
ROLE (reported in~\cite{cole2021multi})  & 1 pos / img                          & -                                          & -               & \textbf{88.2}$^\dag$                         & 69.0$^\dag$  \\ \noalign{\vskip 0.5ex}\hdashline\noalign{\vskip 0.6ex}
Assume negative (AN)                     & 1 pos / img                          & 45.7                                       & {50.2}          & 87.1                                         & 66.9         \\
ROLE (our training schedule)                & 1 pos / img                          & 45.0                                       & 51.9            & 87.8                                         & 69.9        \\\noalign{\vskip 0.5ex}\hdashline\noalign{\vskip 0.6ex}
EN + consistency loss (CL)                    & 1 pos / img                          & 47.0                                       & \textbf{54.3}   & 87.6                                         & 71.6       \\
EN + spatial consistency (SCL)                 & 1 pos / img                          & \textbf{50.4}                              & {54.0}          & 88.0                                         & \textbf{72.1}        \\
\bottomrule
\end{tabular}
}
\end{table*} 
\subsubsection{Implementation details.}
Given $448{\times}448$ inputs, the network outputs $14{\times}14$ score maps. 
Score heatmaps are stored with size $28{\times}28$ in 16-bit floating point format.
We use CL and SCL in combination with EN according to \cref{eq:final_loss}. 
We anneal the SCL weight linearly from $\gamma = 0$ to $\gamma = 1$ in the first 5 epochs of the training, in order to wait for reliable heatmaps. We use the EMA momentum $\mu=0.8$. 
We set the expected number of positives to $K=2.9$ for MS-COCO, and $K=1.5$ for Pascal VOC, based on validation set statistics (see \cref{suppl:dataset_statistics}).

\subsubsection{Results.}

\Cref{tab:results} compares our method to other baselines and related work~\cite{cole2021multi} on Pascal VOC 2012 and MS-COCO 2014 datasets. 
We include results with and without pretraining the ImageNet single-label classification backbone.
The results show that despite its simplicity, the CL is a competitive method: it performs better than AN and performs generally on-par or better than the more sophisticated ROLE method~\cite{cole2021multi}. The SCL further improves upon the results of CL thanks to localized self-supervision.

\subsection{Analysis and ablation}
\label{sec:abl}

Ablation experiments are performed on MS-COCO, with the same setup as in \cref{sec:setup}; we report the best results on the validation split.

\subsubsection{Spatial heatmaps.}
\begin{figure}[tb]
    \centering
    \fullwidthcopyrightimage{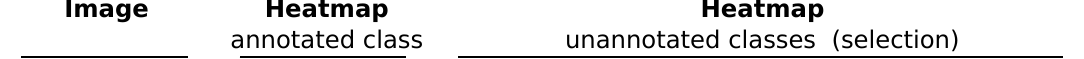}{\phantom{Header}}
    \vspace{0.3em}
\fullwidthcopyrightimage{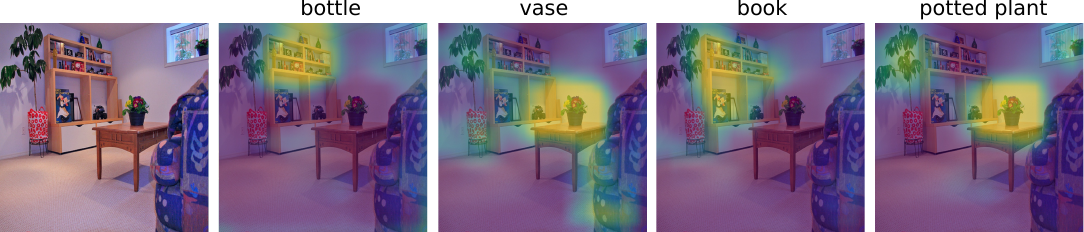}{\href{https://www.flickr.com/photos/waferboard/5484402832/}{Photo by waferboard} \href{https://creativecommons.org/licenses/by/2.0/}{\ccLogo \ccAttribution}}\vspace{0.3em}
    \fullwidthcopyrightimage{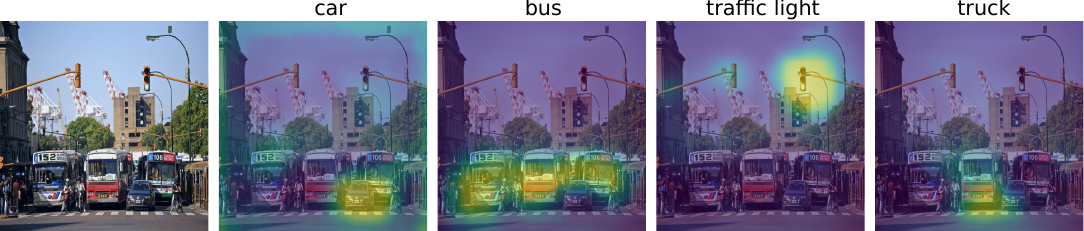}{\href{https://www.flickr.com/photos/proimos/6476201279/}{Photo by 
proimos} \licenseCCBy
    }
\fullwidthcopyrightimage{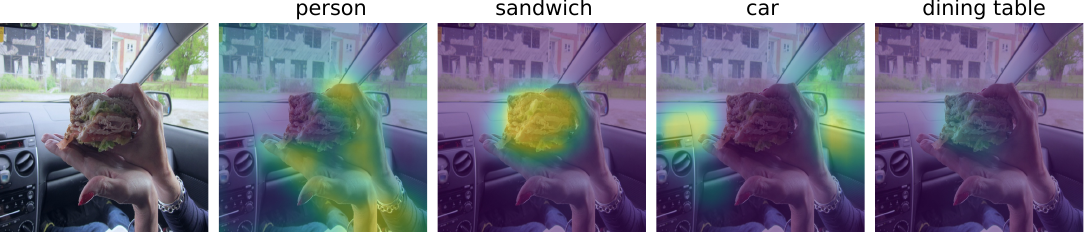}{\href{https://www.flickr.com/photos/infrogmation/6865155270/}{Photo by inforgmation} \href{https://creativecommons.org/licenses/by/2.0/}{\ccLogo \ccAttribution }}
    \caption{Examples of spatial heatmaps produced by ResNet-50 on MS-COCO, in the last training epoch, with ImageNet pretraining (best viewed in color).}
    \label{fig:example_heatmaps}
\end{figure}

\begin{figure}[tb]
    \centering
    \fullwidthcopyrightimage{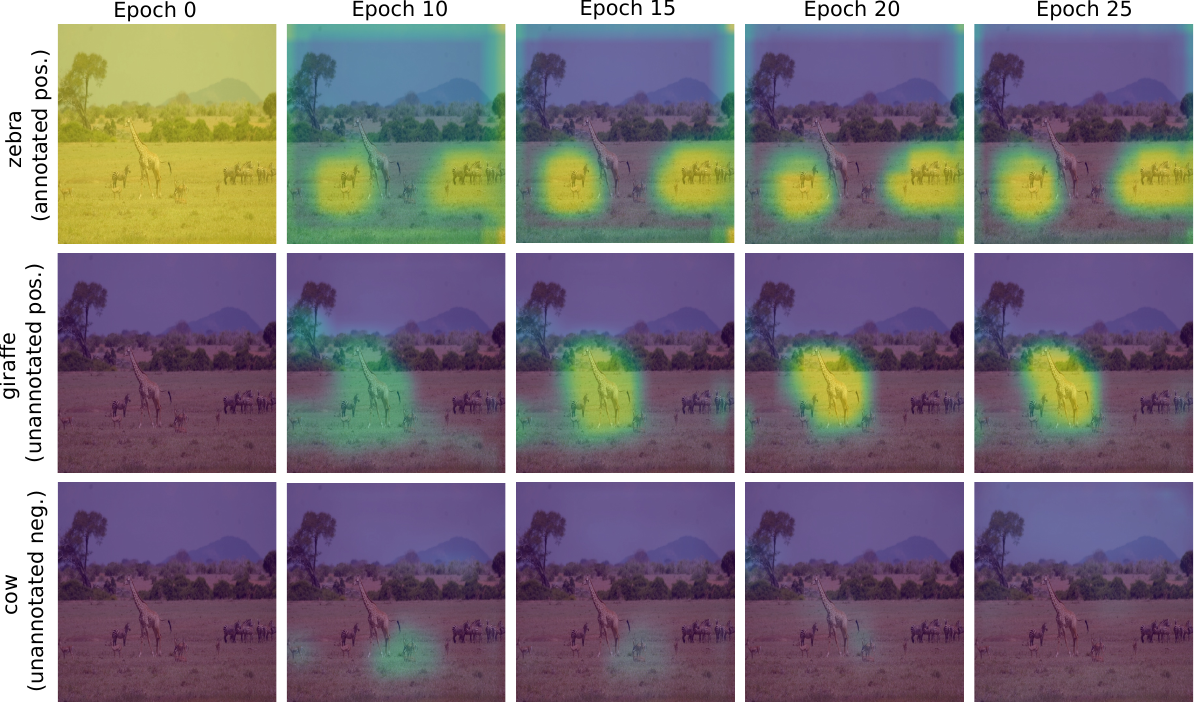}{\href{https://www.flickr.com/photos/lathos/3953904506/}{Photo by S.\ Cozens} \href{https://creativecommons.org/licenses/by-nc/2.0/}{\ccLogo \ccAttribution \ccNonCommercial}}
\caption{Progress of running-average heatmaps during training for an annotated positive class, unannotated positive class and negative class (best viewed in color).}
    \label{fig:running_average_heatmaps}
\end{figure}

\begin{figure}[tb]
    \centering
    \fullwidthcopyrightimage{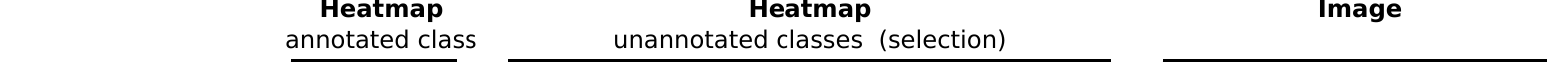}{\phantom{Header}} 
    \vspace{0.3em}
    \fullwidthcopyrightimage{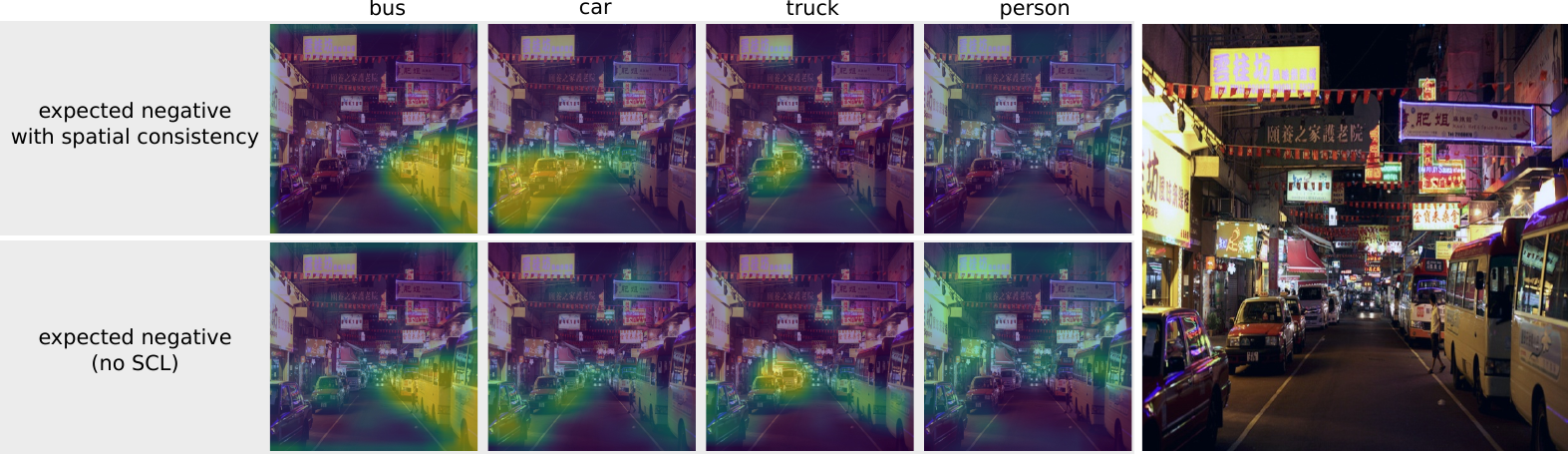}{\href{https://www.flickr.com/photos/scalino/8073309985/}{Photo by 
scalino} \href{https://creativecommons.org/licenses/by-nc/2.0/}{\ccLogo \ccAttribution \ccNonCommercial}}\vspace{0.3em}
    \fullwidthcopyrightimage{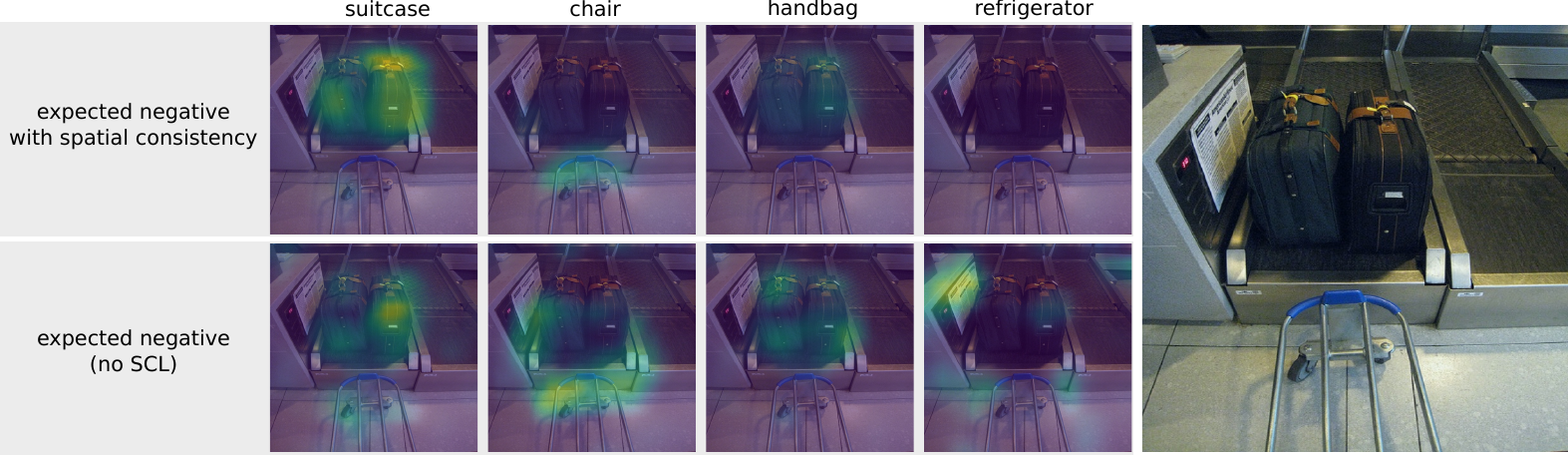}{\href{https://www.flickr.com/photos/blmurch/3124562761/}{Photo by blmurch}\href{https://creativecommons.org/licenses/by/2.0/}{\ccLogo \ccAttribution}
    }
    \caption{Comparison of heatmaps generated in the final training epoch with and without spatial consistency loss.}
    \label{fig:comparison_sc}
\end{figure}

Some qualitative examples of spatial heatmaps are given in \cref{fig:example_heatmaps}. 
We show heatmaps for the positive annotated class, as well as selected heatmaps for unannotated classes. 
The heatmaps exhibit localization of many objects in the image absent from the single-label ground truth. 
\Cref{fig:running_average_heatmaps} shows the progress over the training. 
Comparing the heatmaps with and without $\mathcal{L}_\text{SCL}$ (setting $\gamma{=}0$) in \cref{fig:comparison_sc}, we see that the SCL helps to more precisely localize objects, and avoids false predictions for negative classes. 
We include uncurated heatmaps in \cref{suppl:uncurated_heatmaps} to show that these observations hold in general.

\subsubsection{Bias towards single-positive predictions.}
\label{sec:biasANSP}
\begin{figure}[tb]
    \centering
    \includegraphics[width=0.9\linewidth]{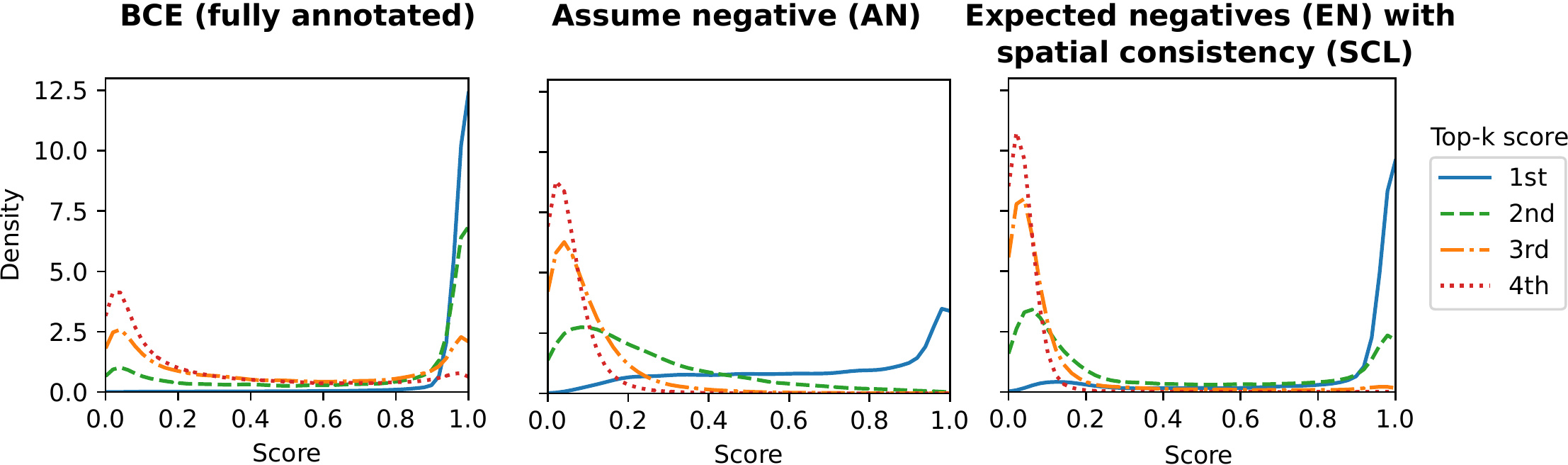}
    \caption{Score distribution over all MS-COCO validation images, for 1st, 2nd, 3rd and 4th highest predicted scores per image. 
    The BCE method is a fully annotated baseline. 
    Training with AN and a single-positive label leads to a bias towards single positive predictions. 
    With EN and SCL, the network more confidently predicts multiple positives.}
    \label{fig:analysis_bias}
\end{figure}

\Cref{fig:analysis_bias} shows the distributions of the top-$4$ scores over all validation images. In contrast to the fully annotated baseline, the single-positive dataset in combination with AN loss leads to low-scoring predictions. 
SCL with EN loss (\cref{eq:final_loss}) reduces the amount of false negative labels and leads to a distribution more akin to the fully annotated case. 

\paragraph{Avoiding the single-positive bias.} In \cref{tab:abl_ignore}, we compare strategies to avoid bias towards single-positive predictions. 
The EN loss in \cref{eq:loss_ignore} leads to ignoring the expected positive samples; we also compare with using them as additional positives in the supervision, using the expected positive loss
\begin{equation} 
\label{eq:loss_expected_positive}
    \mathcal{L}_{\text{EP}}(\mathbf{f}_n) =  -\frac{1}{L} \sum^{L}_{i=1}\, [z_{ni}=1 \lor \hat{z}^t_{ni}=1] \log(f_{ni}) +
    [\hat{z}^t_{ni}=0]  \log(1-f_{ni}) \ .
\end{equation}
We find $\mathcal{L}_\text{EP}$ to perform poorly; we believe incorrect expected-positives disturb the training progress by introducing concept drift. We also compare the EN loss with the expected positive regression loss $\mathcal{L}_\text{EPR}$ of \cite{cole2021multi}, which regresses the sum of the predicted probabilities towards the estimated number of positives $K$.
We see that $\mathcal{L}_\text{EPR}$ performs no better than AN in combination with (S)CL. 
Generally, $\mathcal{L}_\text{EN}$ in combination with $\mathcal{L}_\text{CL}$ or $\mathcal{L}_\text{SCL}$ performs best among competing methods.

\begin{table}[tb!]
\caption{Methods to avoid single-positive bias (MS-COCO \emph{val} split).}
\label{tab:abl_ignore}
\centering
\sisetup{
            detect-all,
            table-number-alignment = center,
            table-figures-integer = 2,
            table-figures-decimal = 1,
}
\begin{tabular}{@{}llS@{}}
\toprule
\textbf{Method}  & \textbf{Loss} & \textbf{mAP} \\ \midrule
assume negative & $\mathcal{L}_\text{AN}$ & 67.4\\
\midrule
assume negative with CL & $\mathcal{L}_\text{AN} + \mathcal{L}_\text{CL}$ & 69.5  \\
expected negatives with CL & $\mathcal{L}_\text{EN} + \mathcal{L}_\text{CL}$   & \bfseries 72.0 \\
expected positives and negatives with CL & $\mathcal{L}_\text{EP} + \mathcal{L}_\text{CL}$   & 64.1 \\
expected positive regression~\cite{cole2021multi} with CL & $\mathcal{L}_\text{EPR}$~(from~\cite{cole2021multi}) + $\mathcal{L}_{CL}$   & 71.0 \\
\midrule
assume negative with SCL & $\mathcal{L}_\text{AN} + \mathcal{L}_\text{SCL}$ & 70.5  \\
expected negatives with SCL & $\mathcal{L}_\text{EN} + \mathcal{L}_\text{SCL}$   & \bfseries 72.5 \\
expected positives and negatives with SCL & $\mathcal{L}_\text{EP} + \mathcal{L}_\text{SCL}$   & 66.0 \\
expected positive regression~\cite{cole2021multi} with SCL & $\mathcal{L}_\text{EPR}$~(from~\cite{cole2021multi}) + $\mathcal{L}_{SCL}$   & 70.7 \\

\bottomrule
\end{tabular}
\end{table}

\begin{figure}[tb]
\centering
\subfloat[{momentum ${\mu}$}]{\includegraphics[width=0.32\linewidth]{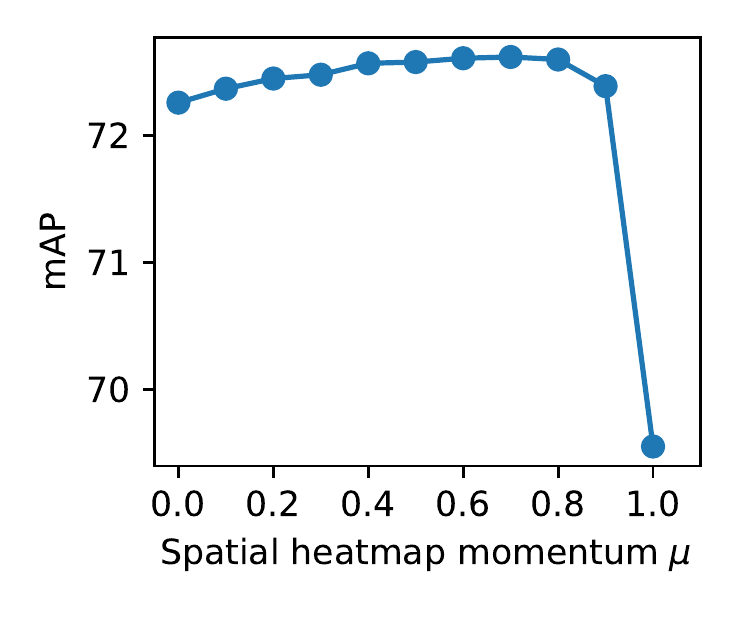}\label{fig:abl_momentum}}
\hfil
\subfloat[hyperparameter $K$]{\includegraphics[width=0.32\linewidth]{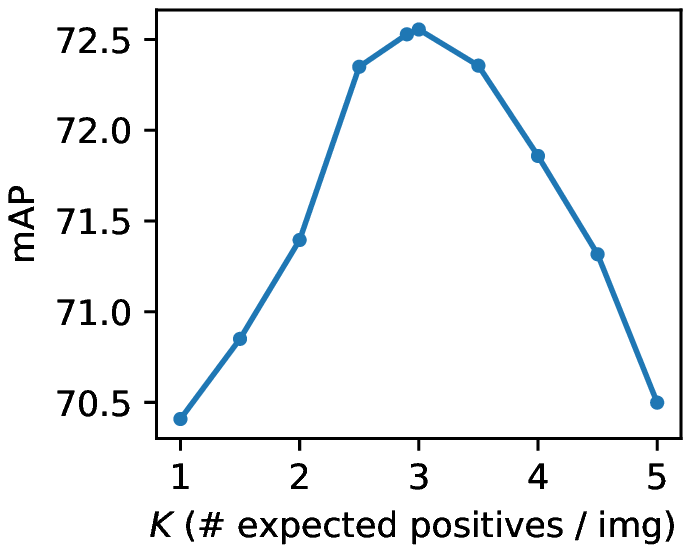}\label{fig:abl_k}} \hfil
\subfloat[annotated positives/img]{\includegraphics[width=0.32\linewidth]{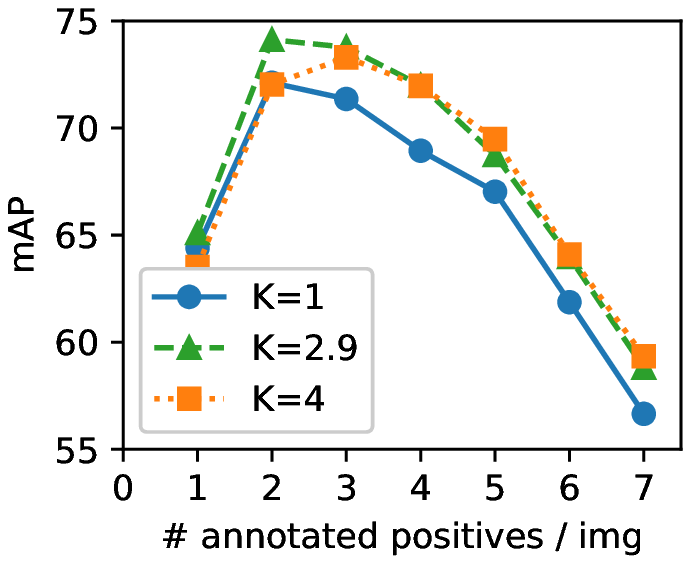} \label{fig:abl_annot_k}}
\caption{Ablations on MS-COCO \textit{val} set with ImageNet-pretrained ResNet-50.}
\end{figure}

\subsubsection{EMA momentum parameter.} \Cref{fig:abl_momentum} compares the validation mAP for values of $\mu$. With $\mu{=}1.0$, heatmaps are not updated by the predictions. 
On the validation set, the value we use in our experiments $\mu=0.8$ corresponds to an optimum between updating the heatmaps and building accurate object localizations.

\subsubsection{Hyperparameter $K$.}  \Cref{fig:abl_k} explores different values for the hyperparameter $K$. For $K{=}2.9$, being determined based on the validation set statistics, our method achieves optimal performance, showing that this prior on the number of positives is well used by the method. 
\Cref{fig:abl_annot_k} compares values of $K$ when restricting the evaluation to images containing $1,2,\dots,7$ true positive labels. 
We see that $K$ tunes the tendency of the classifier to predict more or less positives.

\subsection{Multi-label classification on ImageNet-1K \label{sec:imagenet}}
We apply our method to train a multi-label classifier on ImageNet-1K~\cite{deng_imagenet_2009}, for which multi-label ground truth is not available. 
This single-label dataset has 1.2 million training and 50K validation images. 
As in \cref{sec:setup}, we use a ResNet-50 network pretrained on ImageNet. 
We compare the accuracies obtained when finetuning with AN loss (\cref{eq:anloss}), and EN loss combined with CL or SCL (\cref{eq:final_loss}). 
We use an Adam optimizer~\cite{KingmaB14Adam} with weight decay $10^{-4}$.
The linear classification layer is trained for 5 epochs with learning rate $10^{-4}$ before finetuning the whole network for 25 epochs with cosine learning rate decay. 
We use the standard crop and flip augmentations from~\cite{he2016deep_resnet}. 
We use $224{\times}224$ inputs, leading to score maps of size $7{\times}7$ and heatmaps of size $14{\times}14$ in the SCL. 
To limit the memory usage, we only keep heatmaps for the 10 top-scoring classes after the warmup stage in the SCL (implementation details in \cref{suppl:details_heatmaps}).

We report the top-1 validation accuracy on the ImageNet validation set. 
We also use the relabeled multi-label validation set of ReaL~\cite{beyer2020we_real}, which 
contains annotations for 46837 validation images, having $K=1.22$ positive labels per image on average.
On the ReaL set, we report the top-$1$ accuracy~\cite{beyer2020we_real}\begin{equation}
    \text{top-1}_\text{ReaL} = \frac{1}{N} \sum_{n=1}^N \left[ \argmax{(  \mathbf{f}_n)} \in \{i\,|\,y_{ni} = 1\}\right]\eqsp,
\end{equation}
as well as the 
mean average precision (mAP), and subsets of images having $k = \{1,2,3,4+\}$ labels.
We report all metrics at the end of the finetuning. 

The results are detailed in \cref{tab:imagenet}.
We observe that the finetuning with AN already improves the single-label top-$1$ accuracy of the network, as observed by previous work~\cite{Wightman2021ResNetSB}, as well as multi-label metrics, with a significant boost in mAP. We observe further improvement in the multi-label metrics when adding the CL and SCL losses.
We note that these methods bring the most improvements over AN when looking at the mAP over images with $k=1$ or $k=2$ labels, which constitute $96\%$ of the validation set. 
This is to be expected given the value of the hyperparameter $K=1.2$ for this dataset, which favors images with $1$ or $2$ labels over images with more labels.

\begin{table}[ht]
\centering
    \caption{We finetune 
    ResNet-50 with AN, consistency loss (CL) or spatial consistency loss (SCL). 
    We report top-1 validation accuracy on ImageNet-val (single-label) and on ReaL (multi-label); as well as
mean average precision (mAP) on ReaL. 
    mAP is reported on all images (k = all), or on subsets of images with $k=1, 2, 3, 4+$ annotated labels.}
\label{tab:imagenet}
\sisetup{
            detect-all,
            table-number-alignment = center,
            table-figures-integer = 2,
            table-figures-decimal = 1,
}
\begin{tabular}{@{}lccccccc@{}}
\toprule
        & \multirow{2.4}{*}{\shortstack{top-1\\IN-val}} & \multirow{2.4}{*}{\shortstack{top-1\\ReaL}} & \multicolumn{5}{c}{mAP ReaL}                 \\  \cmidrule{4-8}
                           &   &  & k = all & k = 1 & k = 2 & k = 3 & k $\geq$ 4 \\ \midrule
Num.\ samples & 50,000 & 46,837 & 46,837 & 39,394 & 5,408 & 1,319 & 716 \\ 
\midrule ResNet-50    &  76.1  &  83.0   &   66.3      &  70.6     &  53.0     &  36.1   & \bfseries 22.5      \\
ResNet-50 + AN &  76.9 & 83.1 & 81.4 & 88.0 & 60.0 & \bfseries 36.8 & 21.8 \\ ResNet-50 + EN with CL      & \bfseries 77.1 & \bfseries 83.4 & 81.7 & 88.4 & 60.5 & 36.6 & 21.7 \\
ResNet-50 + EN with SCL &  76.9 & \bfseries 83.4 & \bfseries 82.2 & \bfseries 88.8 & \bfseries 61.4 & \bfseries 36.8 & 21.1 \\
\bottomrule
\end{tabular}
\end{table} 
\subsection{Limitations of the method}
Spatial heatmaps use $2 N L W^2$ bytes of memory, which is around $\SI{16}{\giga\byte}$ for MS-COCO ($N=112\text{K}$, $L=81$, $W=28$). 
For larger datasets, memory constraints can be alleviated by keeping top-$k$ heatmaps after pretraining as we do in \cref{sec:imagenet}, or by offloading the heatmaps to disk with asynchronous I/O. 

Like \cite{cole2021multi} our experiments use an oracle value of the number of expected positives per image $K$ set using statistics from annotated samples. 
This value is dependent on the data collection procedure of the dataset: for instance, ImageNet mostly contains images with one object, whereas MS-COCO images contain many objects. 
Therefore, some calibration of this value is to be expected depending on the dataset and of the properties desired from the classifier. 
 \section{Conclusion}

We studied the problem of training a multi-label classifier using only a single-positive label per image, improving the accuracy using consistency and spatial consistency losses. 
In addition, we showed that standard training strategies result in a bias towards negative predictions and proposed a method to build a set of expected-positive labels, which are not penalized in the training loss.

While we have focused our efforts on the ubiquitous single-positive labeled setting, our works can be naturally extended to partial annotation settings, incorporating multiple positive or negative annotated labels into the loss. 
Besides image crops, other commonly used data-augmentations such as affine transformations or masking correspond to spatially equivariant transformations of the objects localizations in the image, and could be similarly leveraged to enforce consistency of the neural network's feature maps across training epochs. 
Finally, we note that an extension of our approach may also be beneficial in other data modalities which also make use of data augmentations similar to random cropping or masking, such as word deletion in text classification \cite{wei-zou-2019-eda}, or frequency masking with audio data \cite{park2019specaugment}. 
 
\clearpage

\section*{Acknowledgements}
Thomas Verelst and Tinne Tuytelaars acknowledge support from the Flanders AI Research Program.

\section*{Image Attributions}
{\Crefname{figure}{Fig.}{Figs.}
\begin{description}
\item \Cref{fig:splash,fig:running_average_heatmaps}: ``DSC\_9578.jpg'' by Simon Cozens available at \url{https://flickr.com/photos/lathos/3953904506/} under \href{https://creativecommons.org/licenses/by-nc/2.0/}{CC BY-NC 2.0} license
\item \Cref{fig:example_heatmaps}, row 1: ``Downstairs Living Room North East'' by waferboard available at \url{https://flickr.com/photos/waferboard/5484402832/} under \href{https://creativecommons.org/licenses/by/2.0/}{CC BY 2.0} license
\item \Cref{fig:example_heatmaps}, row 2: ``At the Lights, Buenos Aires'' by Alex Proimos available at \url{https://flickr.com/photos/proimos/6476201279/} under \href{https://creativecommons.org/licenses/by-nc/2.0/}{CC BY-NC 2.0} license
\item \Cref{fig:example_heatmaps}, row 3: ``Pt a la Hache Mch 2012 Turkey Sammich'' by Infrogmation of New Orleans available at \url{https://flickr.com/photos/infrogmation/6865155270/} under \href{https://creativecommons.org/licenses/by/2.0/}{CC BY 2.0} license
\item \Cref{fig:suppl_ex_1}, rows 1--2: Photo by julie available at \url{https://flickr.com/photos/roosterfarm/3573516590/} under \href{https://creativecommons.org/licenses/by/2.0/}{CC BY 2.0} license
\item \Cref{fig:suppl_ex_1}, rows 3--4: ``Horse show'' by tanakawho available at \url{https://flickr.com/photos/28481088@N00/2883102207/} under \href{https://creativecommons.org/licenses/by-nc/2.0/}{CC BY-NC 2.0} license
\item \Cref{fig:suppl_ex_1}, rows 5--6: ``Giraffe, Giraffa camelopardalis play fighting at Marakele National Park, Limpopo, South Africa'' by Derek Keats available at \url{https://flickr.com/photos/dkeats/8352611405/} under \href{https://creativecommons.org/licenses/by/2.0/}{CC BY 2.0} license
\item \Cref{fig:suppl_ex_1}, rows 7--8: ``March 22 - 31, 2010'' by osseous available at \url{https://flickr.com/photos/osseous/4530436084/} under \href{https://creativecommons.org/licenses/by/2.0/}{CC BY 2.0} license
\item \Cref{fig:suppl_ex_2}, rows 1--2: ``Birthday ... 10th Dec 1993'' by srv007 available at \url{https://flickr.com/photos/savidgefamily/6809404644/} under \href{https://creativecommons.org/licenses/by-nc/2.0/}{CC BY-NC 2.0} license
\item \Cref{fig:suppl_ex_2}, rows 3--4: ``Fuel stop'' by shirokazan available at \url{https://flickr.com/photos/shirokazan/8033724167/} under \href{https://creativecommons.org/licenses/by/2.0/}{CC BY 2.0} license
\item \Cref{fig:suppl_ex_2}, rows 5--6: ``Frontier Kite Fly Festival 157'' by Michael Kappel available at \url{https://flickr.com/photos/m-i-k-e/2561521549/} under \href{https://creativecommons.org/licenses/by-nc/2.0/}{CC BY-NC 2.0} license
\item \Cref{fig:suppl_ex_2}, rows 7--8: ``USACE division visit to Europe District coincides with German Fasching celebrations'' by U.S. Army Corps of Engineers Europe District available at \url{https://flickr.com/photos/europedistrict/6886413159/} under \href{https://creativecommons.org/licenses/by/2.0/}{CC BY 2.0} license

\end{description}
}

\bibliographystyle{splncs04}
\bibliography{references}

\pagestyle{style2}
\clearpage
\appendix
\counterwithin{figure}{section}
\counterwithin{table}{section}
\section{Data-augmentation settings\label{suppl:dataaug}}
We use the following data-augmentation pipeline during trainings:
\paragraph{MS-COCO 2014 and Pascal VOC 2012}
\begin{description}
\item[Train]
\begin{itemize}
    \item[] \,
    \item Resize to square image of resolution $672{\times}672$
    \item Random square crop with cropped area uniformly varying between 0.25 and 1 (torchvision~\cite{NEURIPS2019_9015} RandomResizedCrop implementation), resized to $448{\times}448$ 
    \item Random horizontal flip
    \end{itemize}
\item[Test]
\begin{itemize}
    \item[] \,
    \item Resize to square image of size $448{\times}448$
\end{itemize}
\end{description}

\paragraph{ImageNet-1k}
\begin{description}
\item[Train]
\begin{itemize}
    \item[] \,
    \item Random square crop with cropped area uniformly varying between 0.08 and 1 and aspect ratio between 3/4 and 4/3 (torchvision~\cite{NEURIPS2019_9015} RandomResizedCrop implementation with default arguments, same as~\cite{he2016deep_resnet}), resized to size $224{\times}224$
    \item Random horizontal flip
    \end{itemize}
\item[Test]
\begin{itemize}
    \item[] \,
    \item Resize smallest image side to 256
    \item Center crop of $224{\times}224$ pixels
\end{itemize}
\end{description}

\begin{figure}[ht]
    \centering
    \includegraphics[width=0.7\linewidth,draft=false]{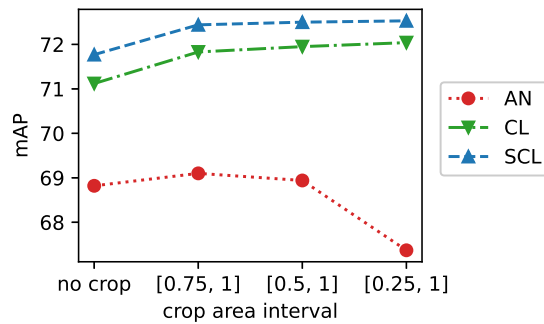}
    \caption{Best MS-COCO validation mAP obtained when training with different data-augmentation crop area. 
    The cropped area size, compared to the full image area, is randomly sampled from the interval uniformly. }
    \label{fig:suppl_crop}
\end{figure}

\subsubsection{Ablation on the crop parameters}
\Cref{fig:suppl_crop} show the accuracies obtained with AN, and CL/SCL (with EN), when varying the random interval for the area of the crop data-augmentation.
We see that CL and SCL are able to benefit from the crop data-augmentation, while AN's performance deteriorates when using that data-augmentation. 
This is consistent with our intuition that the crop data-augmentation can lead to incorrect supervision due to the single annotated objects being possibly partially or entirely cropped out.
Moreover, SCL's improvements over CL are consistent over the different data-augmentation parameters.

\section{Details on heatmaps computation \label{suppl:details_heatmaps}}
We store heatmaps on 2 times the resolution of the feature maps (e.g. input resolution of $448\times448$ results in feature maps of $14 \times 14$ is stored in heatmaps of $28 \times 28$ pixels.). Heatmaps are stored in floating-point 16-bit format. 

For ImageNet-1K~\cite{deng_imagenet_2009} (\cref{sec:imagenet}), we reduce the memory load by only keeping heatmaps for the top-$k$ classes. The selection is based on the per-class EMA scores $\mathbf{s}_n^t$ computed as described in \cref{eq:runningestimate}, after the 5 epochs of pretraining the linear layer. 
In our experiments, we select the $10$ highest-scoring classes per image based on ${s}_{ni}^5$. Heatmaps of other classes are assumed to be uniformly $0$ in the SCL. 
Given $1.3$ million training images, heatmaps of $14\times 14$ and $1000$ classes stored in fp16, this optimization reduces the required memory from approximately $\SI{500}{\giga\byte}$ to $\SI{5}{\giga\byte}$.

\section{Uncurated heatmap examples\label{suppl:uncurated_heatmaps}}
\begin{figure}[t]
    \centering
    \fullwidthcopyrightimage{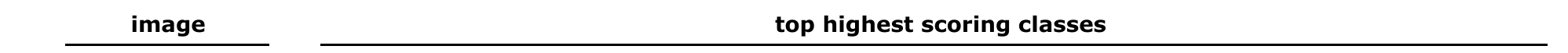}{\phantom{Hj}}\fullwidthcopyrightimage{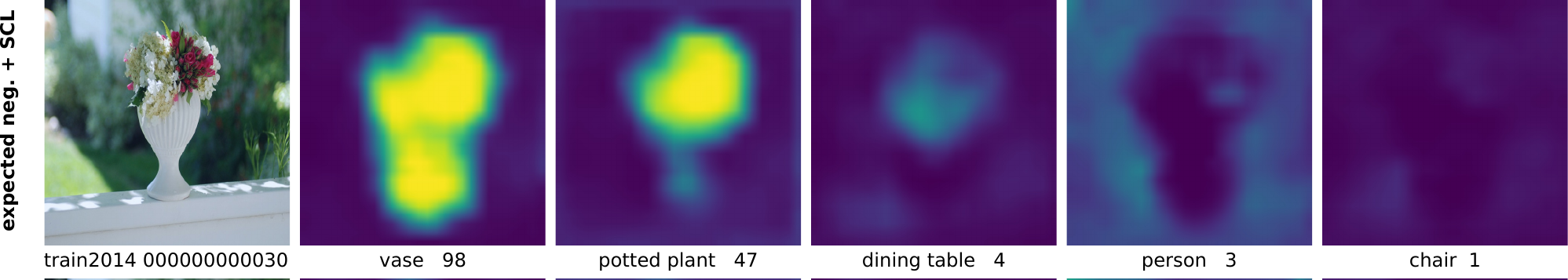}{\phantom{Pj}}\fullwidthcopyrightimage{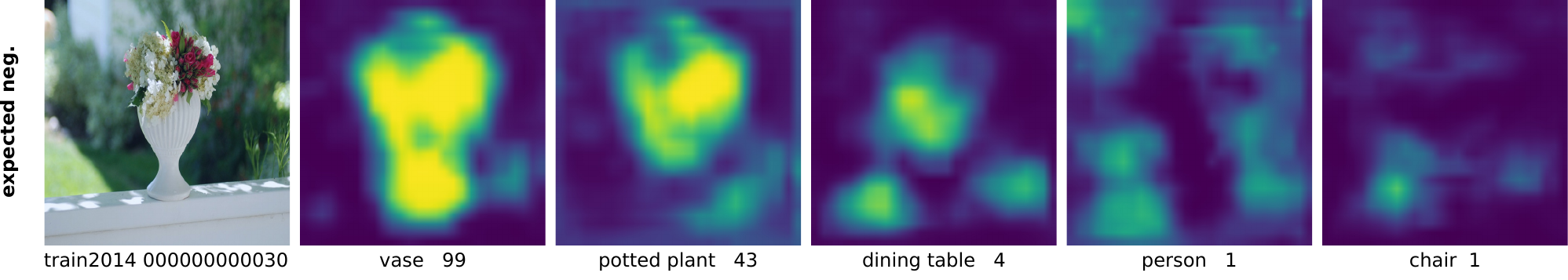}{\href{https://www.flickr.com/photos/roosterfarm/3573516590/}{Photo by 
julie} \href{https://creativecommons.org/licenses/by/2.0/}{\ccLogo \ccAttribution}}\fullwidthcopyrightimage{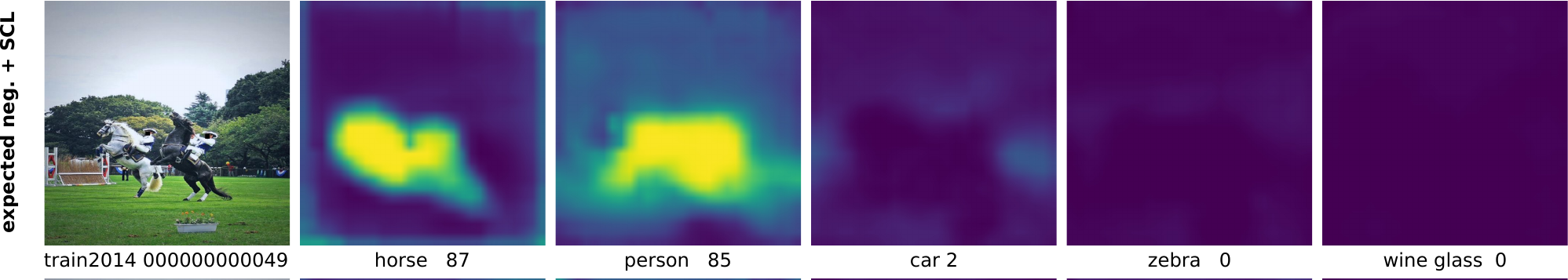}{\phantom{Pj}}\fullwidthcopyrightimage{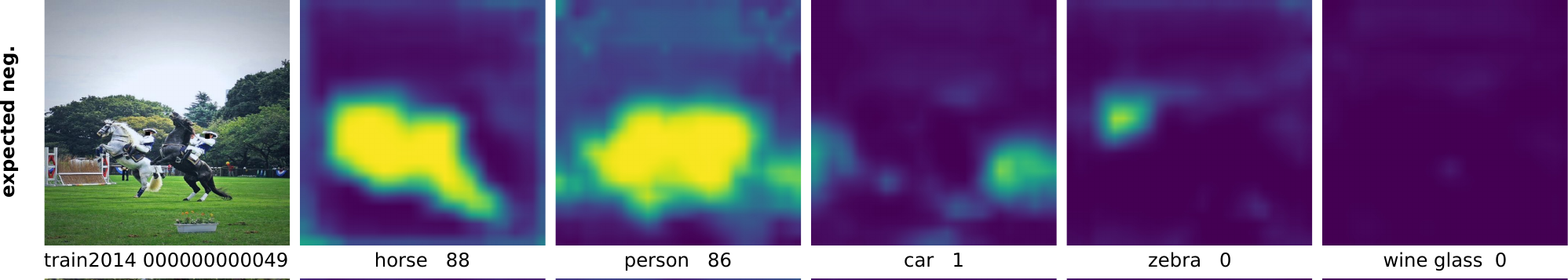}{\href{https://www.flickr.com/photos/28481088@N00/2883102207/}{Photo by tanakawho} \href{https://creativecommons.org/licenses/by-nc/2.0/}{\ccLogo \ccAttribution \ccNonCommercial}
    }\fullwidthcopyrightimage{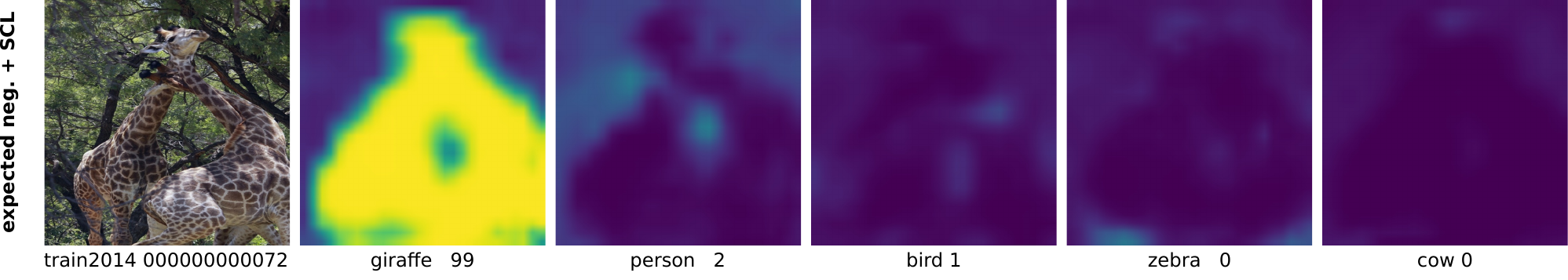}{\phantom{Pj}}\fullwidthcopyrightimage{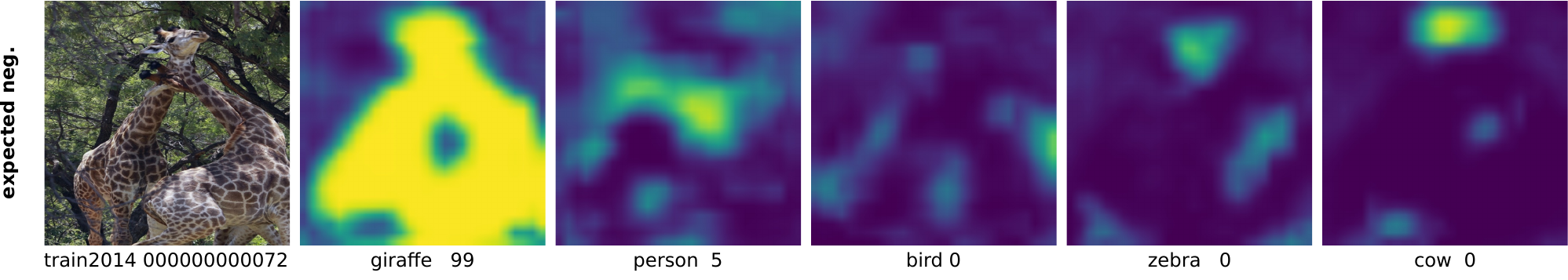}{\href{https://www.flickr.com/photos/dkeats/8352611405/}{Photo by Derek Keats} \href{https://creativecommons.org/licenses/by/2.0/}{\ccLogo \ccAttribution}
    }\fullwidthcopyrightimage{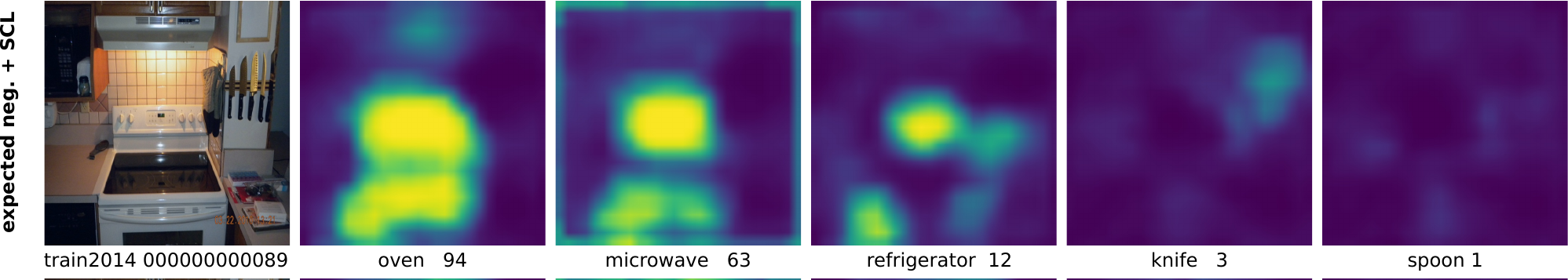}{\phantom{Pj}}\fullwidthcopyrightimage{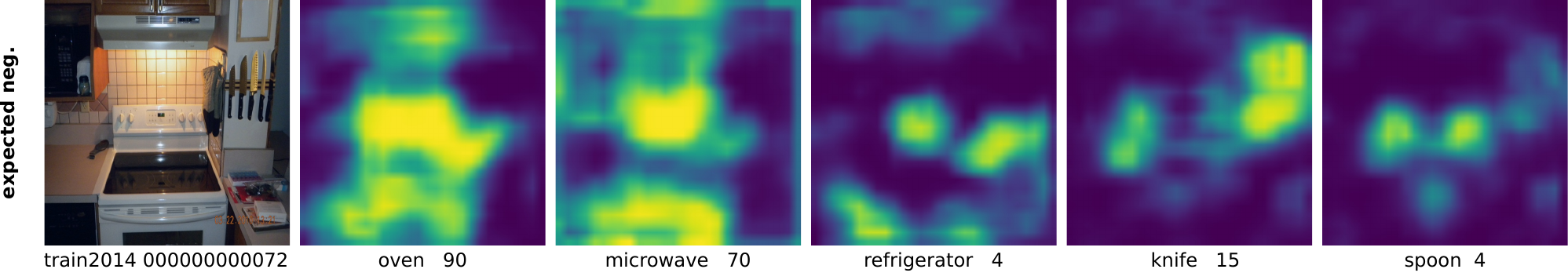}{\href{https://www.flickr.com/photos/osseous/4530436084/}{Photo by osseous} \href{https://creativecommons.org/licenses/by/2.0/}{\ccLogo \ccAttribution}
    }\caption{Heatmaps and scores of the top-$5$ scoring classes in the last epoch training with EN+SCL, along with the corresponding heatmaps for EN alone.}
    \label{fig:suppl_ex_1}
\end{figure}

\begin{figure}[t]
    \centering
    \fullwidthcopyrightimage{fig/suppl/examples/supplExamplesThinheader}{\phantom{Hj}}\fullwidthcopyrightimage{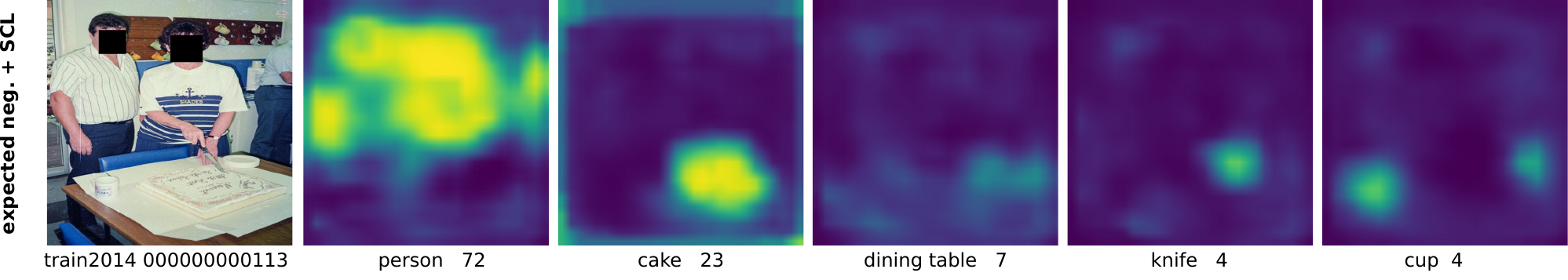}{\phantom{Pj}}\fullwidthcopyrightimage{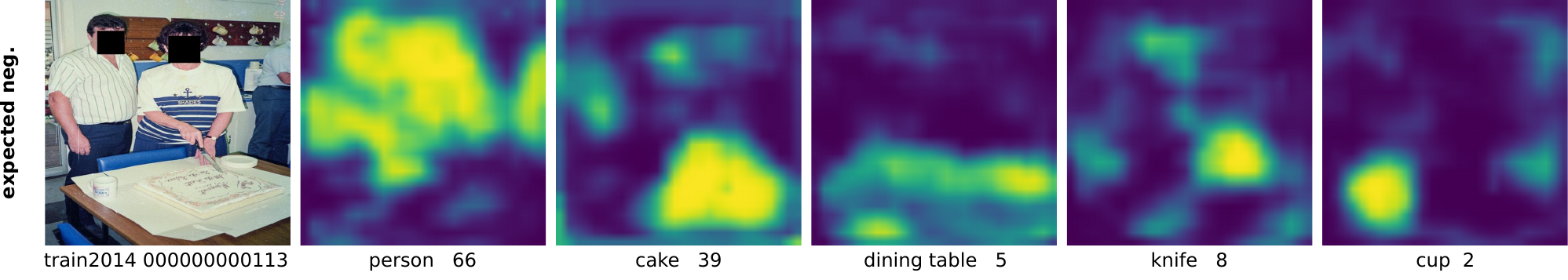}{\href{https://www.flickr.com/photos/savidgefamily/6809404644/}{Photo by 
srv007} \href{https://creativecommons.org/licenses/by-nc/2.0/}{\ccLogo \ccAttribution \ccNonCommercial}}\fullwidthcopyrightimage{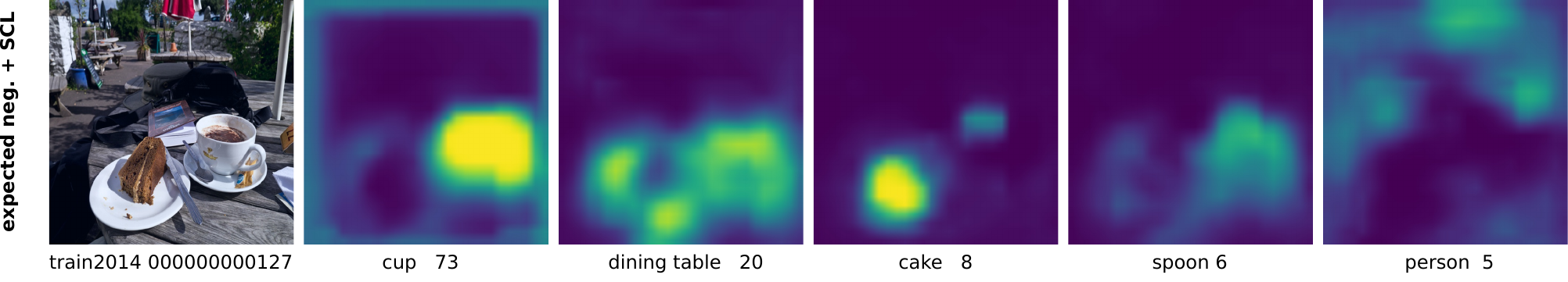}{\phantom{Pj}}\fullwidthcopyrightimage{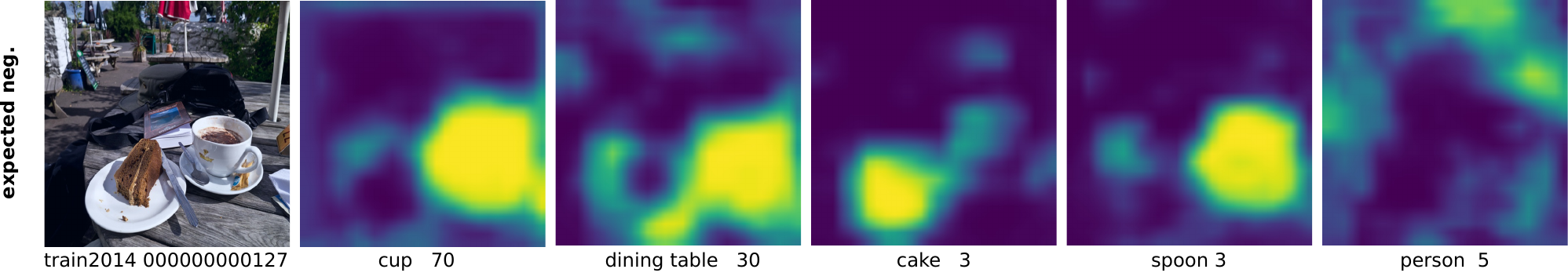}{\href{https://www.flickr.com/photos/shirokazan/8033724167/}{Photo by shirokazan} \href{https://creativecommons.org/licenses/by/2.0/}{\ccLogo \ccAttribution}
    }\fullwidthcopyrightimage{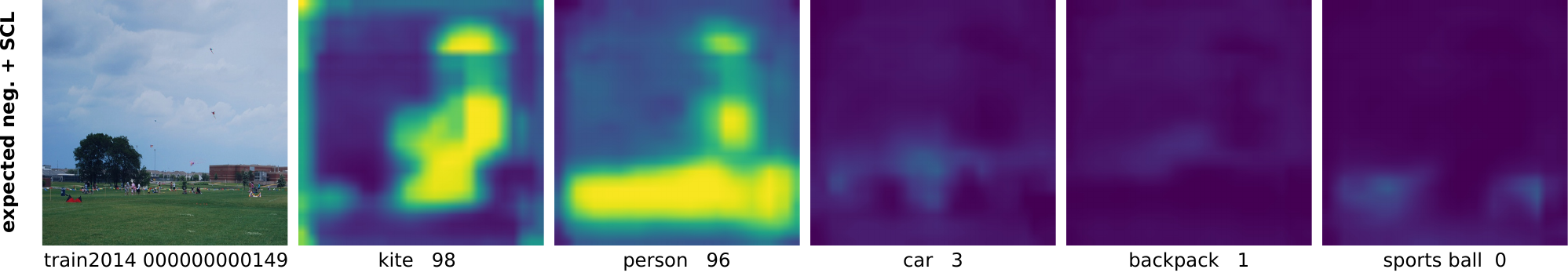}{\phantom{Pj}}\fullwidthcopyrightimage{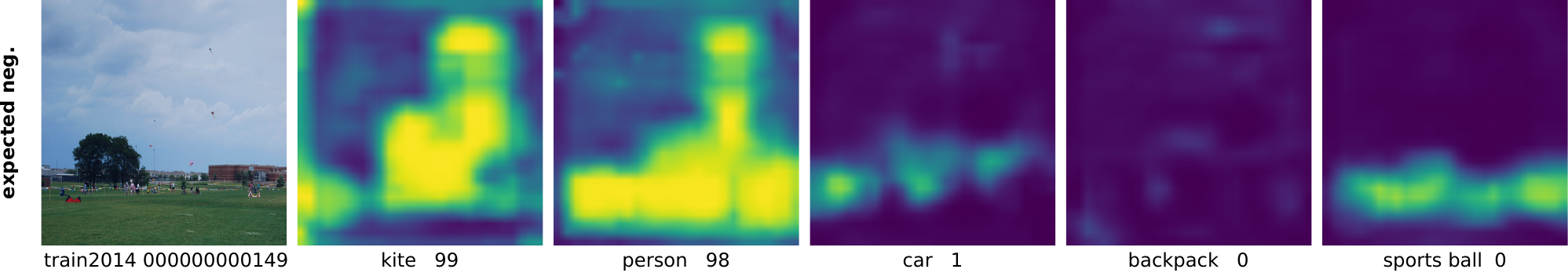}{\makebox[0em][l]{\href{https://www.flickr.com/photos/m-i-k-e/2561521549/}{Photo by Michael Kappel} \href{https://creativecommons.org/licenses/by-nc/2.0/}{\ccLogo \ccAttribution \ccNonCommercial}}
    }\fullwidthcopyrightimage{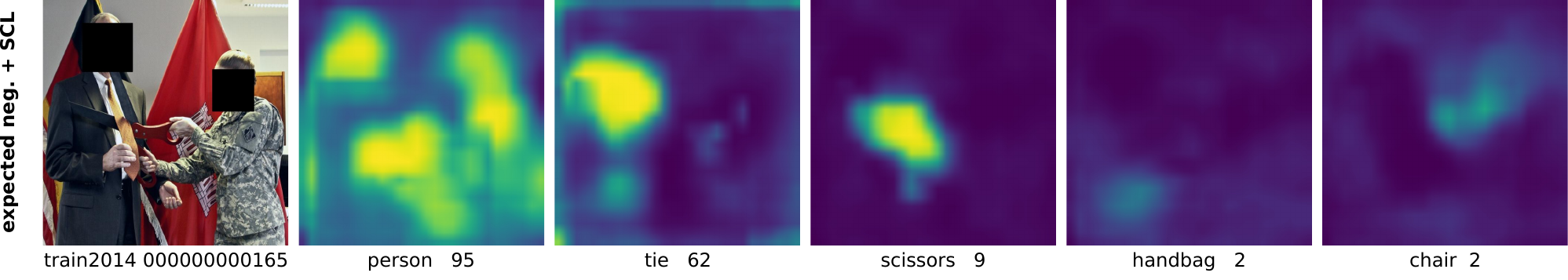}{\phantom{Pj}}\fullwidthcopyrightimage{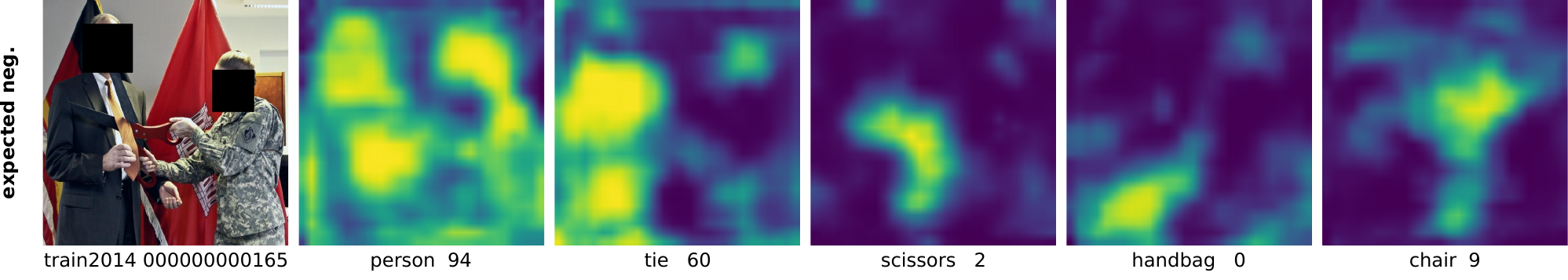}{\makebox[0em][l]{\href{https://www.flickr.com/photos/europedistrict/6886413159/}{Photo by europedistrict} \href{https://creativecommons.org/licenses/by/2.0/}{\ccLogo \ccAttribution}}
    }\caption{
    Heatmaps and scores of the top-$5$ scoring classes in the last epoch training with EN+SCL, along with the corresponding heatmaps for EN alone.}
    \label{fig:suppl_ex_2}
\end{figure}

\Cref{fig:suppl_ex_1,fig:suppl_ex_2} show the heatmaps corresponding to the samples with lowest COCO image id having suitable licenses for reproduction in the paper. 
In agreement with the observations in \cref{sec:abl}, we see that the SCL tends to improve the object localization in the heatmaps, especially when looking at the negative classes which tend to be more present when using the EN alone.

\section{Dataset statistics}
\label{suppl:dataset_statistics}
\Cref{tab:dataset_stats} lists some statistics on the datasets used in the paper, as well as the value of the hyperparameter $K$ computed on the validation set based on these statistics. \Cref{tab:dataset_coco,tab:dataset_pascal} show detailed breakdown of positive annotations per class in the MS-COCO and Pascal datasets using the splits of~\cite{cole2021multi}.

\FloatBarrier
\begin{table}[t]
\centering
\caption{Dataset statistics. 
For COCO and VOC, we use the train/val/test splits from~\cite{cole2021multi}.
For ImageNet-1K we report both the original~\cite{deng_imagenet_2009} and multi-label ReaL~\cite{beyer2020we_real} validation sets. 
$K$ is the average number of positives per image on the validation set.}
\label{tab:dataset_stats}
\adjustbox{max width=\textwidth}{\begin{tabular}{@{}lcccccccc@{}}
\toprule
Dataset                                & Num. classes & \multicolumn{3}{c}{Number of images} & \multicolumn{3}{c}{Number of annotations} & $K$   \\ \cmidrule(lr){3-5} \cmidrule(lr){6-8}
 & &
  \scriptsize train & \scriptsize val & \scriptsize test & \scriptsize train & \scriptsize val & \scriptsize test &
   \\ \midrule
MS-COCO 2014~\cite{lin2014microsoft_coco}   & 80                & 65,665     & 16,416     & 40,137     & 193078         & 47957         & 116592        & 2.9 \\
Pascal VOC 2012~\cite{pascal-voc-2012} & 20                & 4574       & 1143       & 5823       & 6665         & 1143         & 5823        & 1.5 \\
ImageNet-1K~\cite{deng_imagenet_2009} & 1000 & 1,281,167 & 50,000/46,837 & - & 1,281,167 & \multicolumn{1}{c}{50,000/46,837} & - & 1/1.2 \\ \bottomrule
\end{tabular}
}
\end{table}

\begin{table}[tb]
\tiny
\centering
\caption{Annotation statistics on MS-COCO~\cite{lin2014microsoft_coco}.
For each class, we show the total amount of annotations in the original MS-COCO annotations (\emph{total}), as well as the percentage of single-positive annotations selected for that class in the splits of~\cite{cole2021multi}.\label{tab:dataset_coco}}
\def\arraystretch{0.9}
\begin{tabular}{@{}lccccc@{}}
\toprule
\textbf{Class}        & \multicolumn{2}{c}{\textbf{\# train}} & \multicolumn{2}{c}{\textbf{\#  val}}   & \textbf{\# test}  \\
\cmidrule(lr){2-3} \cmidrule(lr){4-5} \cmidrule(lr){6-6}
                        & total     & single-pos     & total & single-pos  & total \\ \midrule
all classes	&193078	&34\%	&47957	&34\%	&116592 \\
\noalign{\vskip 0.5ex}\hdashline\noalign{\vskip 0.6ex}
person	&36192	&34\%	&8982	&34\%	&21634 \\
chair	&7138	&22\%	&1812	&21\%	&4404 \\
car	&6895	&30\%	&1711	&30\%	&4180 \\
dining table	&6701	&21\%	&1677	&21\%	&3960 \\
cup	&5219	&20\%	&1299	&19\%	&3061 \\
bottle	&4790	&20\%	&1178	&21\%	&2912 \\
bowl	&4042	&21\%	&986	&22\%	&2397 \\
handbag	&3927	&23\%	&934	&20\%	&2272 \\
truck	&3447	&33\%	&874	&31\%	&2056 \\
backpack	&3109	&25\%	&815	&25\%	&1832 \\
bench	&3078	&34\%	&766	&35\%	&1961 \\
book	&2994	&22\%	&740	&23\%	&1828 \\
cell phone	&2644	&29\%	&678	&30\%	&1695 \\
sink	&2640	&33\%	&651	&34\%	&1574 \\
tv	&2525	&23\%	&666	&24\%	&1577 \\
couch	&2515	&22\%	&655	&22\%	&1448 \\
clock	&2506	&50\%	&653	&47\%	&1704 \\
potted plant	&2497	&24\%	&587	&23\%	&1540 \\
knife	&2491	&20\%	&606	&19\%	&1410 \\
dog	&2428	&39\%	&613	&39\%	&1521 \\
sports ball	&2401	&30\%	&585	&29\%	&1445 \\
traffic light	&2292	&37\%	&601	&36\%	&1437 \\
cat	&2267	&43\%	&551	&45\%	&1480 \\
bus	&2240	&33\%	&551	&34\%	&1350 \\
umbrella	&2183	&30\%	&566	&32\%	&1393 \\
tie	&2132	&34\%	&535	&34\%	&1288 \\
fork	&2058	&18\%	&479	&17\%	&1173 \\
bed	&2054	&38\%	&485	&39\%	&1292 \\
vase	&2025	&35\%	&505	&36\%	&1200 \\
skateboard	&2021	&40\%	&490	&40\%	&1092 \\
spoon	&2005	&18\%	&488	&21\%	&1189 \\
motorcycle	&1961	&37\%	&481	&38\%	&1219 \\
train	&1958	&58\%	&506	&62\%	&1281 \\
laptop	&1943	&24\%	&532	&24\%	&1232 \\
tennis racket	&1903	&35\%	&465	&37\%	&1193 \\
surfboard	&1876	&44\%	&467	&47\%	&1292 \\
bicycle	&1847	&26\%	&440	&30\%	&1114 \\
toilet	&1842	&58\%	&475	&59\%	&1185 \\
airplane	&1797	&68\%	&446	&69\%	&840 \\
bird	&1784	&64\%	&457	&64\%	&1121 \\
\bottomrule
\end{tabular}
\hfill \begin{tabular}{@{}lccccc@{}}
\toprule
\textbf{Class}        & \multicolumn{2}{c}{\textbf{\# train}} & \multicolumn{2}{c}{\textbf{\#  val}}   & \textbf{\# test}  \\
\cmidrule(lr){2-3} \cmidrule(lr){4-5} \cmidrule(lr){6-6}
               & total      & single-pos     & total     & single-pos     & total \\ \midrule
               \\\noalign{\vskip 0.5ex}\noalign{\vskip 0.6ex}
skis	&1775	&44\%	&434	&43\%	&993 \\
remote	&1750	&25\%	&430	&23\%	&1041 \\
pizza	&1734	&37\%	&468	&37\%	&1117 \\
boat	&1708	&47\%	&390	&43\%	&1048 \\
cake	&1670	&30\%	&410	&29\%	&969 \\
horse	&1668	&52\%	&400	&48\%	&1001 \\
oven	&1584	&26\%	&419	&28\%	&989 \\
baseball glove	&1519	&30\%	&365	&32\%	&845 \\
baseball bat	&1467	&31\%	&337	&30\%	&799 \\
wine glass	&1428	&20\%	&343	&18\%	&872 \\
giraffe	&1426	&80\%	&372	&82\%	&849 \\
sandwich	&1359	&30\%	&286	&31\%	&818 \\
refrigerator	&1344	&27\%	&327	&24\%	&790 \\
banana	&1316	&40\%	&302	&40\%	&728 \\
suitcase	&1313	&34\%	&318	&35\%	&876 \\
kite	&1286	&42\%	&339	&47\%	&727 \\
elephant	&1226	&68\%	&292	&65\%	&714 \\
teddy bear	&1219	&47\%	&291	&47\%	&724 \\
frisbee	&1215	&43\%	&296	&46\%	&757 \\
keyboard	&1161	&21\%	&310	&25\%	&750 \\
cow	&1124	&67\%	&265	&70\%	&666 \\
broccoli	&1080	&41\%	&260	&44\%	&670 \\
zebra	&1065	&86\%	&259	&88\%	&677 \\
mouse	&1008	&23\%	&282	&20\%	&674 \\
orange	&1003	&34\%	&213	&32\%	&568 \\
stop sign	&969	&53\%	&245	&52\%	&589 \\
carrot	&968	&31\%	&218	&35\%	&578 \\
fire hydrant	&954	&52\%	&251	&47\%	&592 \\
apple	&942	&28\%	&229	&31\%	&491 \\
snowboard	&936	&41\%	&234	&42\%	&533 \\
donut	&865	&41\%	&197	&40\%	&523 \\
sheep	&856	&73\%	&249	&75\%	&489 \\
microwave	&853	&23\%	&236	&25\%	&512 \\
hot dog	&661	&38\%	&160	&41\%	&452 \\
toothbrush	&570	&36\%	&130	&49\%	&341 \\
scissors	&535	&44\%	&138	&42\%	&302 \\
bear	&531	&88\%	&137	&86\%	&341 \\
parking meter	&395	&42\%	&86	&50\%	&261 \\
toaster	&125	&28\%	&26	&23\%	&74 \\
hair drier	&103	&27\%	&25	&28\%	&70 \\
\bottomrule
\end{tabular}

\vspace{0.4cm}
\caption{Annotation statistics on Pascal VOC 2012~\cite{pascal-voc-2012}.
For each class, we show the total amount of annotations in the original MS-COCO annotations (\emph{total}), as well as the percentage of single-positive annotations selected for that class in the splits of~\cite{cole2021multi}.\label{tab:dataset_pascal}}
\begin{tabular}{@{}lccccc@{}}
\toprule
\textbf{Class}        & \multicolumn{2}{c}{\textbf{\# train}} & \multicolumn{2}{c}{\textbf{\#  val}}   & \textbf{\# test}  \\
\cmidrule(lr){2-3} \cmidrule(lr){4-5} \cmidrule(lr){6-6}
                        & total     & single-pos     & total & single-pos  & total \\ \midrule
all classes	&6665	&68\%	&1666	&68\%	&8351 \\
\noalign{\vskip 0.5ex}\hdashline\noalign{\vskip 0.6ex}
person	&1584	&59\%	&410	&66\%	&2093 \\
dog	&504	&83\%	&128	&82\%	&654 \\
car	&474	&68\%	&116	&60\%	&571 \\
chair	&459	&49\%	&107	&39\%	&553 \\
cat	&436	&90\%	&103	&87\%	&541 \\
bird	&310	&93\%	&85	&98\%	&370 \\
bottle	&294	&51\%	&71	&52\%	&341 \\
aeroplane	&264	&95\%	&63	&95\%	&343 \\
tvmonitor	&233	&61\%	&57	&63\%	&285 \\
diningtable	&221	&45\%	&48	&43\%	&269 \\
\bottomrule
\end{tabular}
\quad \begin{tabular}{@{}lccccc@{}}
\toprule
\textbf{Class}        & \multicolumn{2}{c}{\textbf{\# train}} & \multicolumn{2}{c}{\textbf{\#  val}}   & \textbf{\# test}  \\
\cmidrule(lr){2-3} \cmidrule(lr){4-5} \cmidrule(lr){6-6}
               & total      & single-pos     & total     & single-pos     & total \\ \midrule
               \\\noalign{\vskip 0.5ex}\noalign{\vskip 0.6ex}
train	&220	&85\%	&53	&83\%	&271 \\
pottedplant	&214	&47\%	&55	&61\%	&258 \\
boat	&210	&82\%	&50	&72\%	&248 \\
motorbike	&206	&65\%	&59	&50\%	&261 \\
sofa	&201	&53\%	&56	&58\%	&250 \\
bicycle	&200	&64\%	&68	&66\%	&284 \\
horse	&195	&69\%	&42	&66\%	&245 \\
bus	&176	&67\%	&37	&64\%	&208 \\
sheep	&135	&90\%	&36	&86\%	&154 \\
cow	&129	&86\%	&22	&90\%	&152 \\
\bottomrule
\end{tabular}
\end{table} 

 \end{document}